\documentclass[12pt]{elsarticle}

\journal{}

\usepackage{tabularx}
\usepackage{geometry}
\usepackage{multirow}
\usepackage{graphicx}
\usepackage{booktabs}
\usepackage{floatpag}
\usepackage{hyperref}
\usepackage{subcaption}
\usepackage{amsmath}
\usepackage[inline]{enumitem}
\usepackage{pgfplots, pgfplotstable}
\usepackage[UKenglish]{isodate}
\usepackage[section]{placeins}
\usepackage{tikz}
\usetikzlibrary{shapes, positioning, decorations.pathreplacing, fit, patterns, shadows}
\definecolor{train}{HTML}{baed91}
\definecolor{val}{HTML}{b2cefe}
\definecolor{test}{HTML}{fea3aa}

\tikzstyle{tsstep} = [rectangle, draw, minimum size=0.5cm]

\newcolumntype{Y}{>{\centering\arraybackslash}X}

\begin{document}

\begin{frontmatter}

\title{The effects of regularisation on RNN models for time series forecasting: Covid-19 as an example}
\author[uni]{Marcus Carpenter\corref{email}}
\author[uni]{Chunbo Luo}
\author[bt]{Xiao-Si Wang}
\address[uni]{Department of Computer Science, University of Exeter, North Park Road, Exeter EX4 4QF, UK}
\address[bt]{Applied Research, British Telecommunications plc., Adastral Park, Ipswich, UK}
\cortext[email]{mc844@exeter.ac.uk}

\begin{abstract}
Many research papers that propose models to predict the course of the COVID-19 pandemic either use handcrafted statistical models or large neural networks. Even though large neural networks are more powerful than simpler statistical models, they are especially hard to train on small datasets. This paper not only presents a model with grater flexibility than the other proposed neural networks, but also presents a model that is effective on smaller datasets. To improve performance on small data, six regularisation methods were tested. The results show that the GRU combined with 20\% Dropout achieved the lowest RMSE scores. The main finding was that models with less access to data relied more on the regulariser. Applying Dropout to a GRU model trained on only 28 days of data reduced the RMSE by 23\%.
\end{abstract}

%
%


\end{frontmatter}

\section{Introduction}
Naive time series models tend to perform better than complex time series models. The main problem with simpler statistical models is that, even though their performance may be better, they cannot be generalised for multiple locations. To model the COVID-19 pandemic in a new country a new model must be created. Neural networks, on the other hand, are good at generalising a dataset. This means that neural networks can learn from data of one country and interpolate this when making predictions for another location. The problem is that most neural networks models that have been proposed are vary large, deep learning requires larger datasets, and no model seems to take advantage of its generalisation capabilities. Essentially all neural network models proposed are as flexible as the statistical models, but lack their simplicity.

The data provided by the Johns Hopkins University contains multiple  multivariate time series data for a total of 187 different locations. These locations are mostly countries; however, some cruise ships have been included. To make predictions on all available locations using a single model, a multi-output RNN model is proposed. Moreover, regularisation methods were applied and tested on the RNN in an attempt to improve performance on small datasets.

Regularisers add extra penalties during the training process that shrink the weights of the model and force it to learn more abstract functions. A regularised model is more robust to noisy data and less prone to overfitting the training samples. The results show that applying regularisers on the models reduced the RMSE of the predictions, and the models relied more on regularisers when trained on less data. In other words, RNN models trained on larger datasets do not require regularisers and applying them can even harm performance.

The code for the experiments can be found at: 

\url{https://github.com/marcusCarpenter97/COVID-19-forecasting}

\section{Related work}
\label{sec:related_work}
Statistical models were surveyed based on their type or complexity. As for the neural network models, there have been many proposed predicative models. The two main traits evaluated in this survey are the size and complexity of the neural networks, and the model's capability to generalise the data well enough.

Traditional models include mathematical, statistical and simple machine learning models. One major limitation of these, simpler, models is that they must be fitted on individual time series. Therefore, simple models do not have the flexibility of producing forecasts for multiple locations at once. Examples of handcrafted mathematical models include Polynomials \cite{Nandi2020} and Logistic growth \cite{Jain2020}. Compartmental models from epidemiology were also used including SEIR \cite{Bhati2020} and SIR \cite{Ghanbari2020}. More complex statistical models, such as the Gaussian Process Regression \cite{Khan2020} and ARMA \cite{Maleki2020}, have also been successfully applied to forecasting COVID-19. Rustam et al. \cite{Rustam2020} showed that Exponential Smoothing outperformed the Support Vector Machine further proving that simpler models can outperform more complex ones.

The Multilayer Perceptron (MLP) was used for forecasting the COVID-19 infection trends. Wieczorek et al. \cite{Wieczorek2020} trained an eight layer MLP over 3300 epochs to benchmark multiple optimisation algorithms. This is a fairly large model considering it was only capable of producing forecasts for 30 location at a time. Hazarika et al. \cite{Hazarika2020} created a wavelet coupled random vector functional link network (WCRVFL). This model is a type of MLP that integrates the wavelet transform for signal processing, in this case a time series. Their experiments focused only on the top five worst hit countries and their predictions spanned a period of 60 days. However, these 60 days were in the future at the time the article was published, therefore it is impossible to verify the quality of forecasts because the data did no yet exist at the time of writing.

The RNN, and mainly its variant the Long Short Term Memory (LSTM), was used in the literature. Shastri et al. \cite{Shastri2020} focused on the LSTM and its variants. Their experiments showed that the ConvLSTM achieved the lowest Mean Absolute Percentage Error (MAPE) among all other models. The MAPE is not a suitable error metric for a dataset that contain zeros which the Johns Hopkins University data does. Predictions were only made for India and the USA. Zeroual et al. \cite{Zeroual2020} not only experimented on the LSTM but also included the RNN, GRU and the Variational AutoEncoder (VAE). The models were trained for 1000 epochs and made predictions for six countries using a forecast horizon of 17 days. The RMSE was used to measure forecast performance, and the VAE was determined to be the best model.

\section{Methods}
\label{sec:methods}
It is helpful to first mathematically define the model M as

\begin{equation}
    \hat{y}_{t+1:t+n}^l = M(x_{1:t}^l, ID^l, \theta)
    \label{equation:model}
\end{equation}
where, for a location \(l\), \(x_{1:t}^l\) represents the time series data from day 1 to \(t\), \(ID^l\) is a location's unique identifier and \(\theta\) stands for the model's parameters. The model outputs the predictions for each day between days \(t+1\) and \(t+n\) for location \(l\). The values of \(t\) and \(n\) represent the last day in the training data and the forecast horizon, respectively.

The data used in the experiments was collected from the Johns Hopkins University GitHub repository \cite{Dong2020}. This dataset contains 187 main locations. These locations tend to be countries, although some cruise ships where included in the data. Because the data contains large values, training the neural networks on the original data would produce undesirable results. To address this, the data was standardised.

\subsection{Model architecture}
\begin{figure}
    \centering
    
    \tikzset{%
      cascaded/.style = {%
        general shadow = {%
          shadow scale = 1,
          shadow xshift = -1ex,
          shadow yshift = 1ex,
          draw,
          thick,
          fill = white},
        general shadow = {%
          shadow scale = 1,
          shadow xshift = -.5ex,
          shadow yshift = .5ex,
          draw,
          thick,
          fill = white},
        fill = white, 
        draw,
        thick,
        minimum width = 2cm,
        minimum height = 1cm}}
    \resizebox{\textwidth}{!}{
    \begin{tikzpicture}
        \node (lid) [cascaded] at (0.1, 0) {Location ID};
        \node (tsd) [cascaded] at (0.1, -2) {Time Series};
        
        \node (dns) [circle, draw, thick] at (3.3, 0) {Linear};
        \node (rnn) [rectangle, draw, thick, minimum height=2cm] at (3.3,-2) {RNN encoder};
        
        \node (ctx) [rectangle, draw, thick, minimum height=1cm] at (6.3, -1) {Context};
        
        \node (cds) [rectangle, draw, thick, minimum width=3cm] at (9.4, 0) {Confirmed layer};
        \node (dds) [rectangle, draw, thick, minimum width=3cm] at (9.4, -1) {Deceased layer};
        \node (rds) [rectangle, draw, thick, minimum width=3cm] at (9.4, -2) {Recovered layer};
        
        \node (cts) [cascaded] at (13, 0.5) {Time Series};
        \node (dts) [cascaded] at (13, -1) {Time Series};
        \node (rts) [cascaded] at (13, -2.5) {Time Series};
        
        \draw[->] (lid) -- (dns);
        \draw[->] (tsd) -- (rnn);
        \draw[->] (dns) -- (ctx);
        \draw[->] (rnn) -- (ctx);
        \draw[->] (ctx) -- (cds);
        \draw[->] (ctx) -- (dds);
        \draw[->] (ctx) -- (rds);
        \draw[->] (cds) -- (cts);
        \draw[->] (dds) -- (dts);
        \draw[->] (rds) -- (rts);
        
        \node[draw, rectangle, fit=(lid)(tsd), minimum height=4cm, minimum width=3cm, label=Input] {};
        \node[draw, rectangle, fit=(dns)(cds)(rnn)(rds), minimum height=4.5cm, minimum width=1cm, label=Model] {};
        \node[draw, rectangle, fit=(cts)(rts), minimum height=5cm, minimum width=3cm, label=Output] {};
    \end{tikzpicture}
    }
    \caption{The model takes as input a location ID and a multivariate time series for each country. The ID is processed by a single linear node for its affine transformation, and the time series is processed by the RNN layer. The outputs of the two are concatenated into a context which is then processed by three separate branches of fully connected nodes. Each branch specialises on one of the features in the data.}
    \label{fig:architecture}
\end{figure}
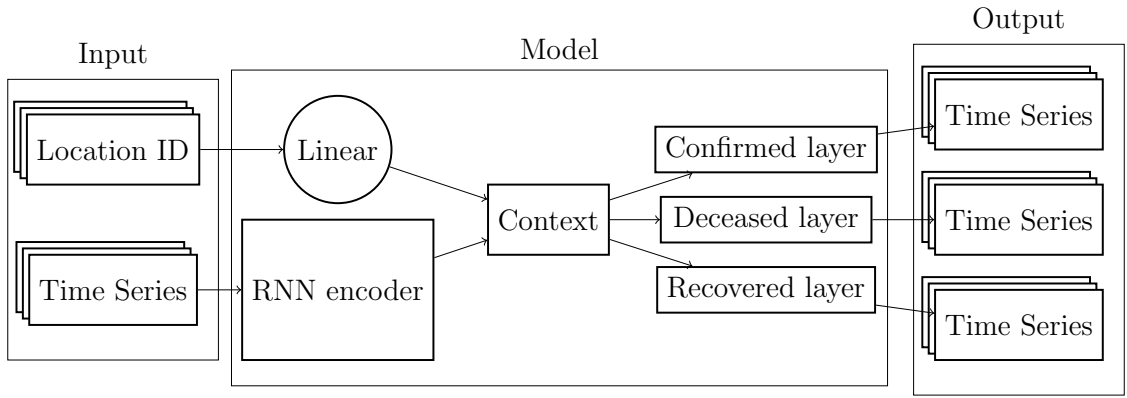

The model uses a multi-output encoder-decoder architecture. Figure \ref{fig:architecture} illustrates the overall architecture of the model.

The encoder takes two inputs: the time series data for a location and its unique identifier. The time series is processed by the RNN layer. The first six characters were extracted from the SHA-256 hash of the location's name and used as the unique identifier for that specific location. This identifier allows the model to better differentiate between different time series; however, Wieczorek et al. stated that this was not necessary for their experiments to succeed \cite{Wieczorek2020}. The identifier was processed by a linear node for its affine transformation. The context is created by concatenating the outputs of the RNN and the linear node.

The decoder is composed of three branches. Each branch is made of a single layer of linear nodes, and the number of nodes is the same as the number of days in the forecast horizon, in this case 28. This branching of the output allows the model to be optimised not only for each of the features but also for each of the days in the forecast horizon.

\subsection{Regularisers}
\label{section:regularisers}
Regularisation methods can be applied to neural networks with the objective of preventing it from overfitting the training data. This allows the model to become more robust to noisy data because of its new generalisation capabilities. Table \ref{table:regularisers} lists the six regularisers explored in the experiments.

The LASSO (L1) and Ridge regression (L2) impose an extra penalty on the model's loss function. This extra penalty to the loss function forces the weights of the neural network to shrink during training, effectively serving as a feature selection \cite{Muthukrishnan2016}. The two regularisers can be combined into what is called the ElasticNet by adding the two penalty factors together \cite{Zou2005}. The main improvement ElasticNet brings is that the L2 method prevents the L1 form shrinking the weights down to zero. A sparse weight matrix in a neural network would lead to many connections becoming ineffective. Equations \ref{equation:l1} and \ref{equation:l2} show the L1 and L2, where \(\lambda\) is the hyperparameter for the penalty factor.

\begin{equation}
    L1 = loss(M(x_{1:t}^l, ID^l, \theta)) + \lambda \sum^{n}_{i=1} |w_{i}|
    \label{equation:l1}
\end{equation}
\begin{equation}
    L2 = loss(M(x_{1:t}^l, ID^l, \theta)) + \lambda \sum^{n}_{i=1} w^{2}_{i}
    \label{equation:l2}
\end{equation}

Dropout, on the other hand, is a regularisation method that purposefully makes the weight matrix sparse by deactivating randomly selected nodes. This, however, only occurs during the training process, and during the testing all nodes are fully functional. The amount of deactivated nodes is selected as a percentage. When Dropout is used, the model is forced to learn a more general function to compensate for the lack of extra nodes \cite{Hinton2012}.

\begin{table}
    \centering
    \begin{tabular}{cccc}
        \hline
        \hline
        Label & L1 & L2 & Dropout \\
        \hline
        No reg & 0 & 0 & 0 \\
        L1 & 0.01 & 0 & 0 \\
        L2 & 0 & 0.01 & 0 \\
        Dropout & 0 & 0 & 0.2 \\
        L1L2 & 0.01 & 0.01 & 0 \\
        All reg & 0.01 & 0.01 & 0.2 \\
        \hline
        \hline
    \end{tabular}
    \caption{The different combination of regularisers used in the experiments. Note that the ElasticNet was labelled L1L2 and No reg represents a standard model without any regularisers applied to it.}
    \label{table:regularisers}
\end{table}

\section{Experiments}
\label{sec:experiments}
All models were trained for 300 epochs using the Adam optimizer with learning rate of 0.001 and the Mean Squared Error (MSE) as the loss function. The output layer used a linear activation function mathematically defined as \(f(x) = x\). The L1 and L2 regularisers were applied on the weights of the RNN encoder which used one, 20 node, layer. Both, L1 and L2, regularisers used \(\lambda = 0.01\) while the dropout percentage was 0.2 or 20\%. Table \ref{table:regularisers} outlines how the regularisers where arranged into six combinations. 

The models were cross-validated. Cross-validation provides a robust statistical evaluation of the model and allows the output for each time period to be visualised. When cross-validating a time series special precautions have to be made due to the temporal nature of the data. To address this, the forward chaining method was used with a forecast horizon of 28 days.

In the experiments the data for each location contained 448 days (04/02/2020 to 27/04/2021). Both the validation and test sets contained 28 days allowing the data to be split into 14 validation folds. The size of the training data grew from 28 to 392 days and not 448 as the last section of the data must be kept for the validation and test sets.

\section{Discussion}
\label{sec:discussion}
After running the experiments, the models were evaluated for model selection. There were fourteen cross-validation folds, six regularisers, two model architectures, and, for each of these, an ensemble of ten models. The best model was selected based on the smallest average RMSE of the model's ensemble. Each model produced a very large output of 187 multivariate time series, and therefore the mean of the RMSE was taken for each of these outputs. It is important to note that, because the RMSE is not scale invariant, this average simply becomes a number to minimise rather than an estimation of people the model missed in its predictions.

The model with the lowest mean and the lowest variance on its RMSE ensemble was selected as the best. Out of all candidates, the best performing model was the GRU combined with 20\% Dropout rate. This model was compared with the standard GRU (no regularisers) to analyse the affects of applying Dropout to a GRU. Table \ref{table:cross-validation} reveals that Dropout had the greatest impact on the first validation fold where it reduced the RMSE by 23\%; Dropout negatively impacted the model on the last two validation folds by up to a 4\% increase in RMSE. In other words, models trained on smaller time series rely more heavily on the regulariser, models trained on medium sized time series do not rely so much on regularisation methods, and models trained on larger time series can experience a degradation in performance if combined with Dropout.

\begin{table}[ht]
    \centering
    \begin{tabular}{rrrrr}
        \hline
        \hline
        Fold & Dates & No reg & Dropout & Change \\
        \hline
         0 & 31/03/2020 to 28/04/2020 &   493.796 &   379.001 & -23\% \\
         1 & 28/04/2020 to 26/05/2020 &   341.281 &   341.321 &   0\% \\
         2 & 26/05/2020 to 23/06/2020 &   410.034 &   399.417 &  -3\% \\
         3 & 23/06/2020 to 21/07/2020 &   614.241 &   590.560 &  -4\% \\
         4 & 21/07/2020 to 18/08/2020 &   711.503 &   706.253 &  -1\% \\
         5 & 18/08/2020 to 15/09/2020 &   644.467 &   646.871 &   0\% \\
         6 & 15/09/2020 to 13/10/2020 &   723.875 &   697.384 &  -4\% \\
         7 & 13/10/2020 to 10/11/2020 &  1355.365 &  1306.728 &  -4\% \\
         8 & 10/11/2020 to 08/12/2020 &  1623.798 &  1515.455 &  -7\% \\
         9 & 08/12/2020 to 05/01/2021 & 10206.108 & 10182.144 &   0\% \\
        10 & 05/01/2021 to 02/02/2021 &  1361.537 &  1353.741 &  -1\% \\
        11 & 02/02/2021 to 02/03/2021 &   930.507 &   906.355 &  -3\% \\
        12 & 02/03/2021 to 30/03/2021 &  1187.182 &  1188.860 &   0\% \\
        13 & 30/03/2021 to 27/04/2021 &  2101.707 &  2178.459 &   4\% \\
        \hline
        \hline
    \end{tabular}
    \caption{The RMSE values for the GRU with no regularisers and the GRU with dropout are presented for all validation folds along with the change in RMSE. Smaller RMSE is better and so is a greater negative change. Positive change indicates that the baseline model performed better.}
    \label{table:cross-validation}
\end{table}

The data can be categorised into four distinct types, these are: smooth, outlier, step, and flat. These categories affect how the model handles its predictions. \emph{Smooth data} is the most predominant type in the dataset. This is a time series with no anomalies which allows the model to predict it with accuracy. \emph{Outlier data} occurs when there is a sudden spike on some of the days. This occurs rarely, and the reason for this is likely to be a mistake in the reporting of the data. The models handle outliers by ignoring them and simply following the trend of the data. \emph{Step data} is defined as having sudden jumps in the recorded cases. Even though this can be a genuine spike in cases, it is likely that the data collection was intermittent. The models tend to produce forecasts that resemble smooth data. Consequently, this can be interpreted as the model having learnt how the number of cases grows and being able to make predictions for days with missing data. \emph{Flat data} is defined by an unchanging number of reported cases either because of a stagnation of the disease or a lack of data. The model produces forecasts as real numbers; therefore, it cannot output a precise constant value. To overcome this problem, the model produced predictions that fluctuate around the constant.

\section{Conclusion}
\label{sec:conclusion}
Many neural network models have been applied to forecasting the trends of COVID-19 infections. However, the fact that deep learning requires large amounts of data and that the time series available of the pandemic is relatively small is often overlooked. A regularised GRU was proposed that is capable of producing accurate forecasts even when trained on a 28 day time series. From the results, it is possible to conclude that when making time series forecasts using an RNN, regularisers can greatly improve performance. However, this is only the case on small datasets and prediction accuracy was found to decrease when regularisers were applied on larger time series. 

\appendix

\section{Plots}
\label{apx:plots}

\begin{figure*}[ht!]
    \centering
    
    \resizebox{0.9\textwidth}{!}{
    \begin{subfigure}{0.5\textwidth}
        \centering
        \includegraphics[scale=0.3]{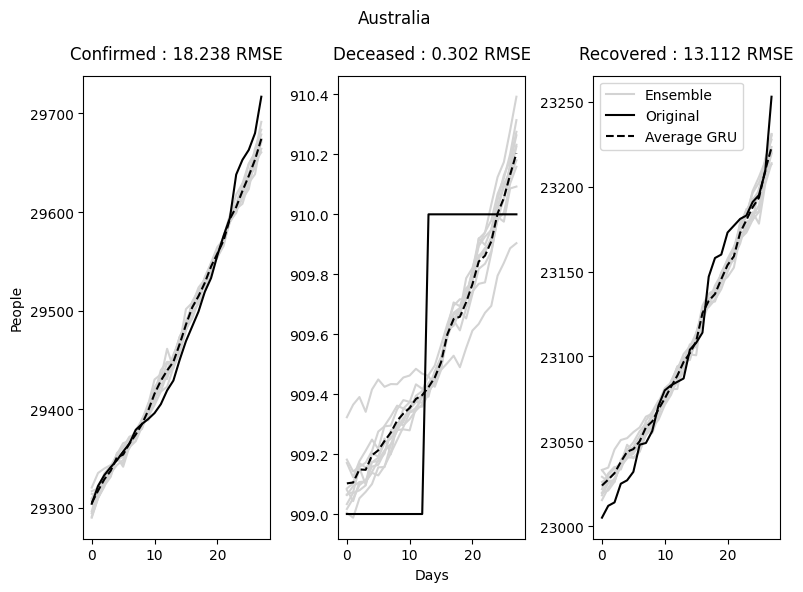}
        \caption{Australia}
    \end{subfigure}%
    \begin{subfigure}{0.5\textwidth}
        \centering
        \includegraphics[scale=0.3]{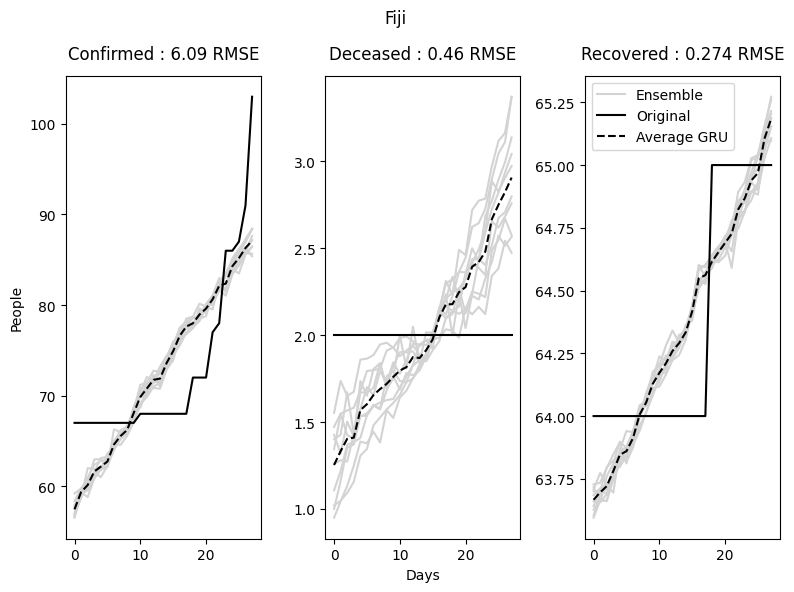}
        \caption{Fiji}
        \label{fig:Fiji}
    \end{subfigure}
    }
    \resizebox{0.5\textwidth}{!}{
    \begin{subfigure}{0.5\textwidth}
        \centering
        \includegraphics[scale=0.3]{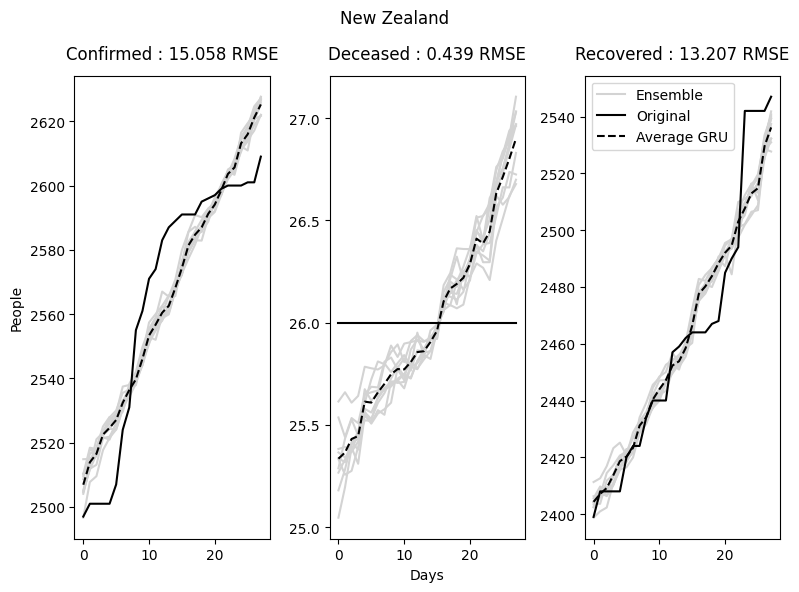}
        \caption{New Zealand}
        \label{fig:New Zealand}
    \end{subfigure}
    }
    \caption{Oceania form 30/03/2021 to 27/04/2021}
    \label{fig:Oceania}
\end{figure*}
\begin{figure*}
    \centering
    
    \resizebox{\textwidth}{!}{
    \begin{subfigure}{0.5\textwidth}
        \centering
        \includegraphics[scale=0.25]{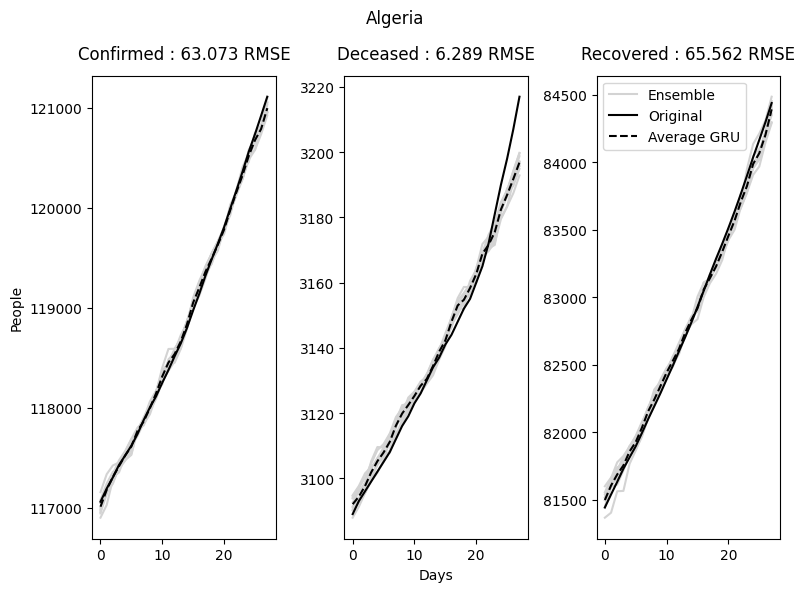}
        \caption{Algeria}
    \end{subfigure}%
    \begin{subfigure}{0.5\textwidth}
        \centering
        \includegraphics[scale=0.25]{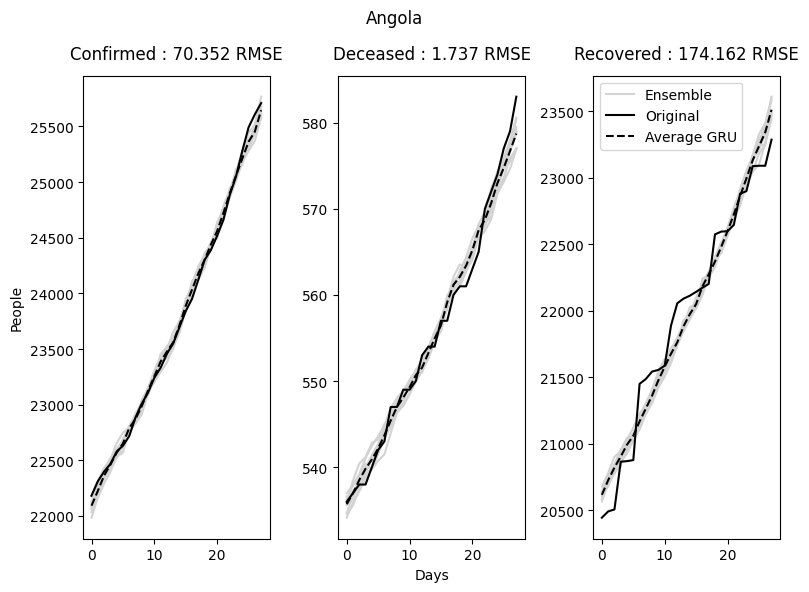}
        \caption{Angola}
    \end{subfigure}
    }
    \resizebox{\textwidth}{!}{
    \begin{subfigure}{0.5\textwidth}
        \centering
        \includegraphics[scale=0.25]{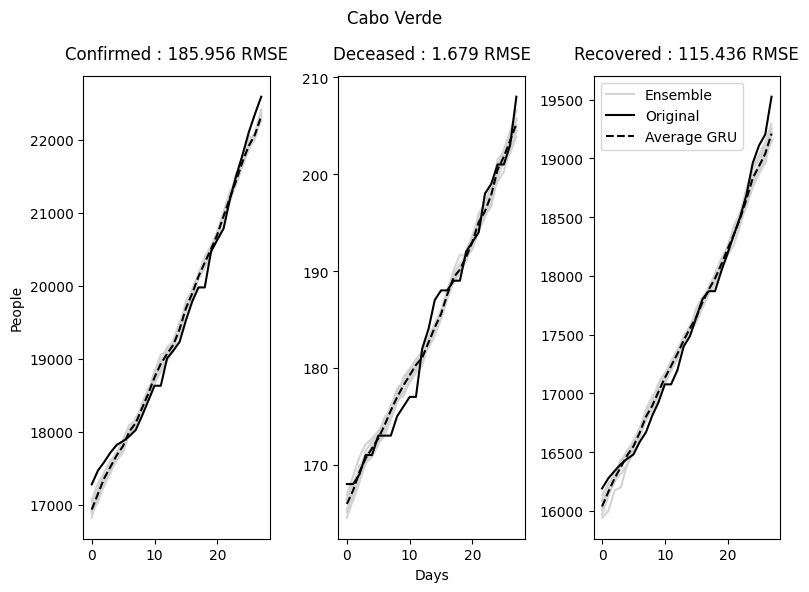}
        \caption{Cabo Verde}
    \end{subfigure}%
    \begin{subfigure}{0.5\textwidth}
        \centering
        \includegraphics[scale=0.25]{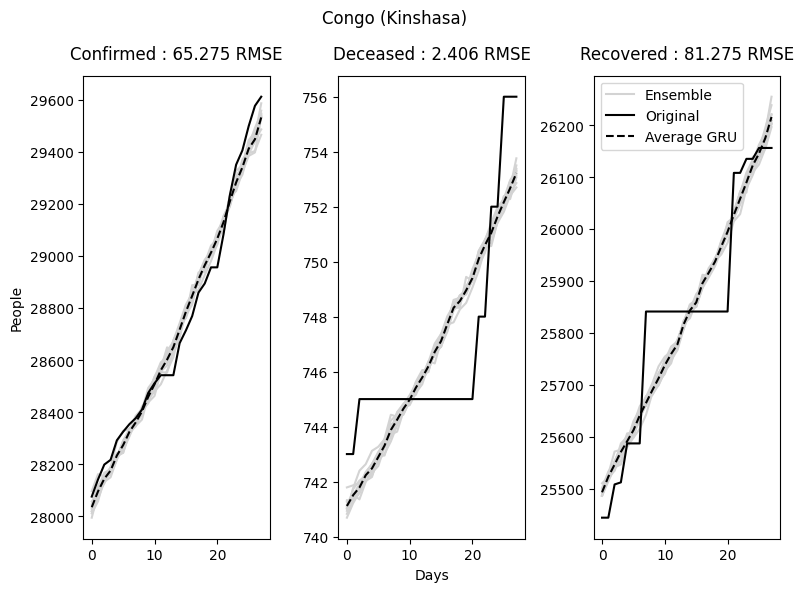}
        \caption{Congo}
    \end{subfigure}
    }
    \resizebox{\textwidth}{!}{
    \begin{subfigure}{0.5\textwidth}
        \centering
        \includegraphics[scale=0.25]{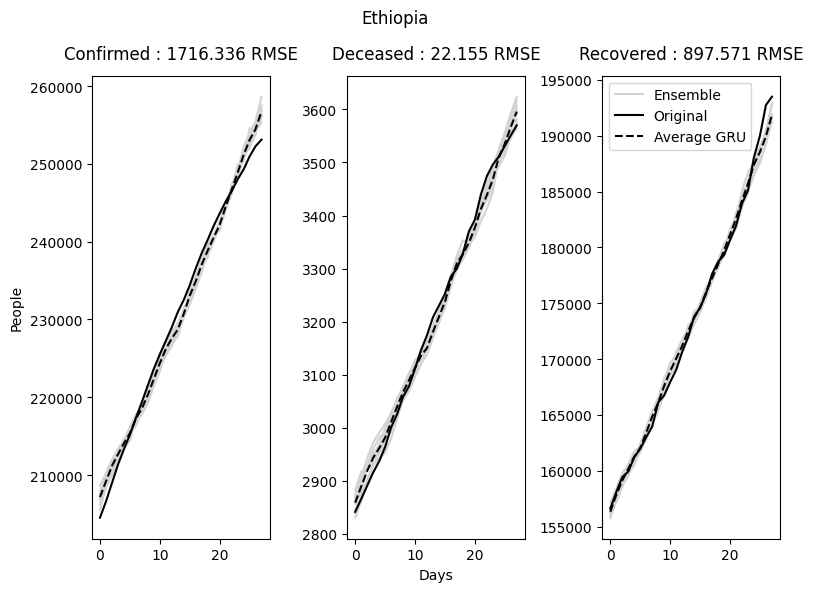}
        \caption{Ethiopia}
    \end{subfigure}%
    \begin{subfigure}{0.5\textwidth}
        \centering
        \includegraphics[scale=0.25]{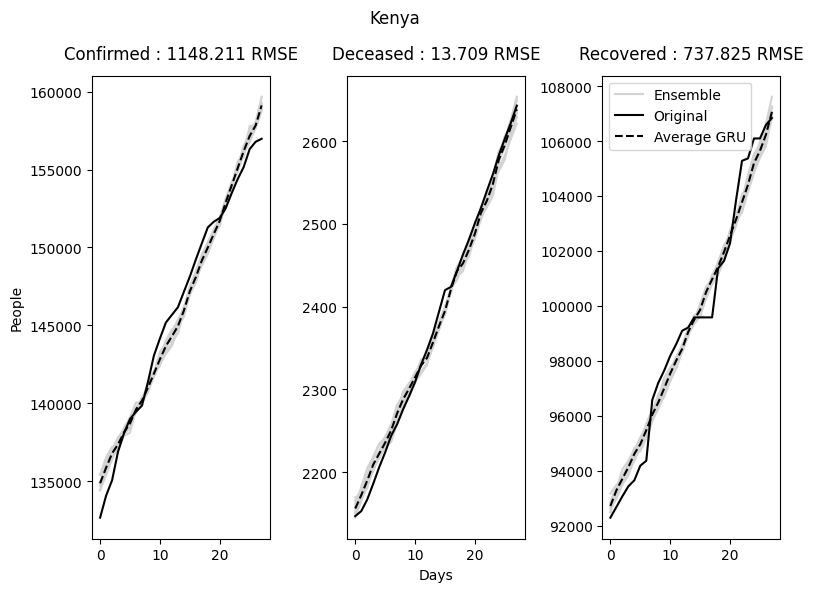}
        \caption{Kenya}
    \end{subfigure}
    }
    \resizebox{\textwidth}{!}{
    \begin{subfigure}{0.5\textwidth}
        \centering
        \includegraphics[scale=0.25]{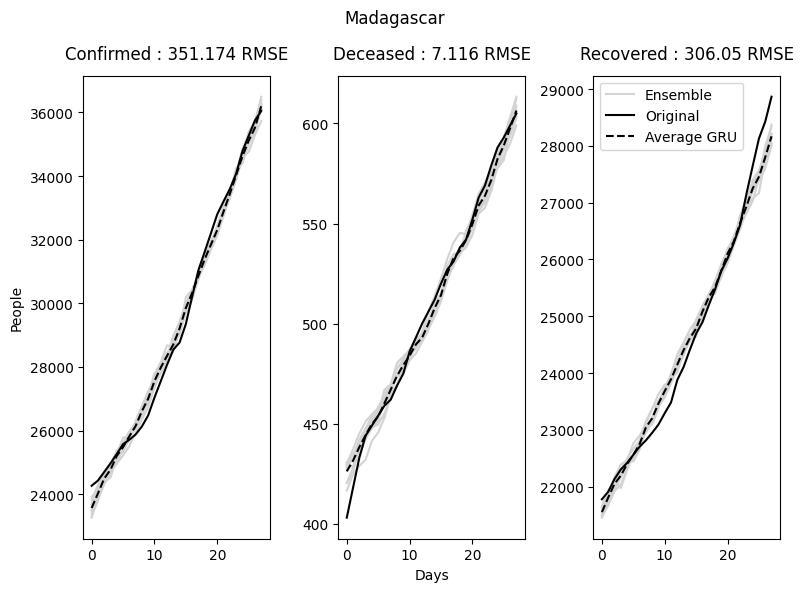}
        \caption{Madagascar}
    \end{subfigure}%
    \begin{subfigure}{0.5\textwidth}
        \centering
        \includegraphics[scale=0.25]{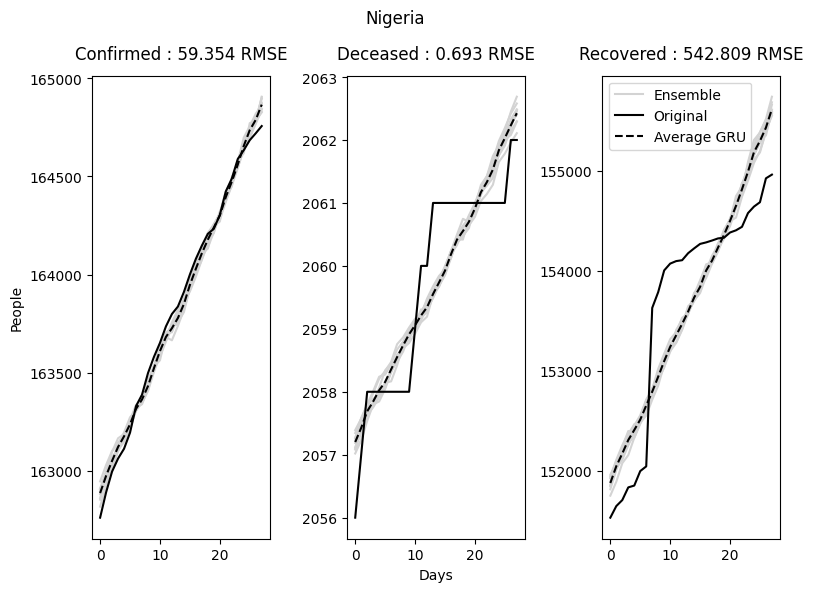}
        \caption{Nigeria}
    \end{subfigure}
    }
    \resizebox{\textwidth}{!}{
    \begin{subfigure}{0.5\textwidth}
        \centering
        \includegraphics[scale=0.25]{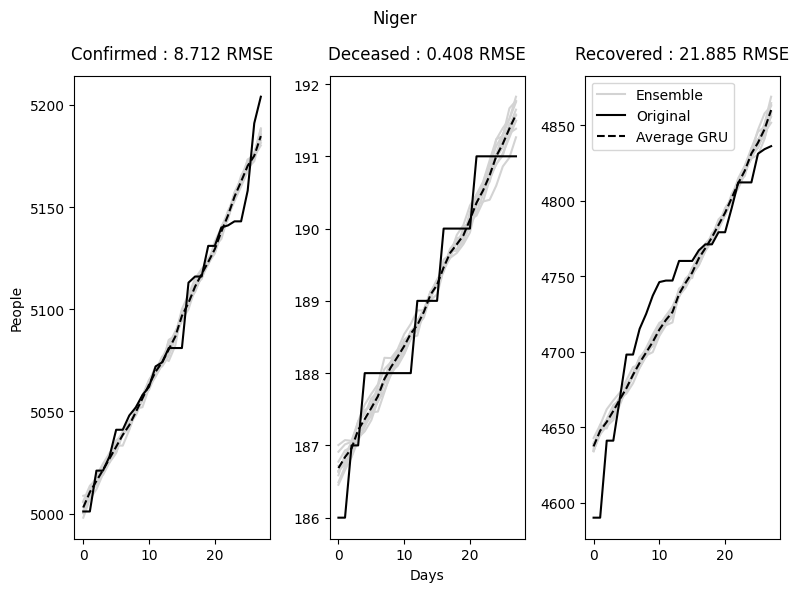}
        \caption{Niger}
        \label{fig:Niger}
    \end{subfigure}%
    \begin{subfigure}{0.5\textwidth}
        \centering
        \includegraphics[scale=0.25]{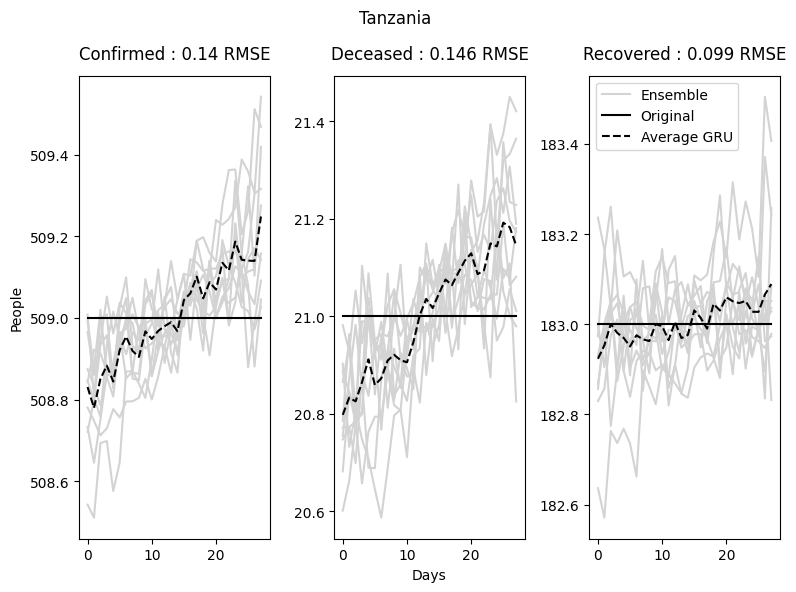}
        \caption{Tanzania}
        \label{fig:Tanzania}
    \end{subfigure}
    }
    \caption{Africa form 30/03/2021 to 27/04/2021}
    \label{fig:Africa}
\end{figure*}
\begin{figure*}
    \centering
    
    \resizebox{\textwidth}{!}{
    \begin{subfigure}{0.5\textwidth}
        \centering
        \includegraphics[scale=0.25]{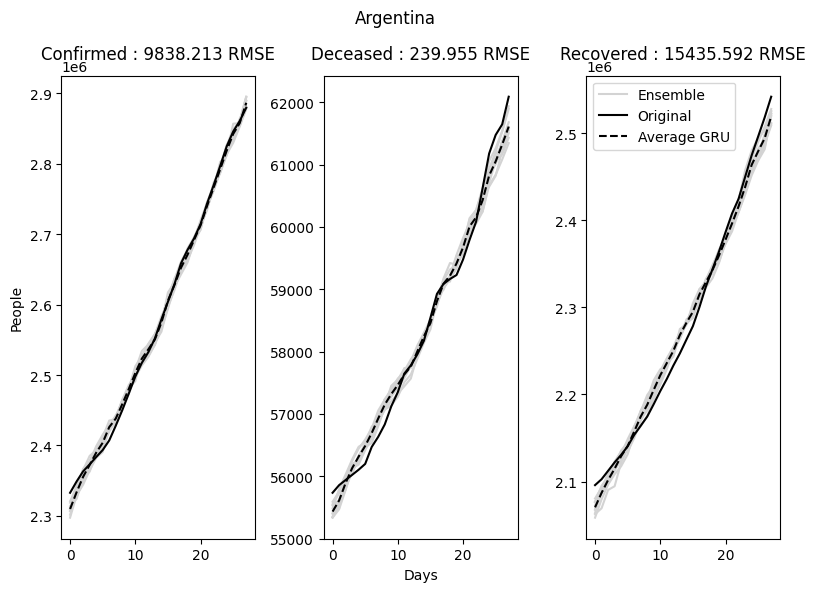}
        \caption{Argentina}
    \end{subfigure}%
    \begin{subfigure}{0.5\textwidth}
        \centering
        \includegraphics[scale=0.25]{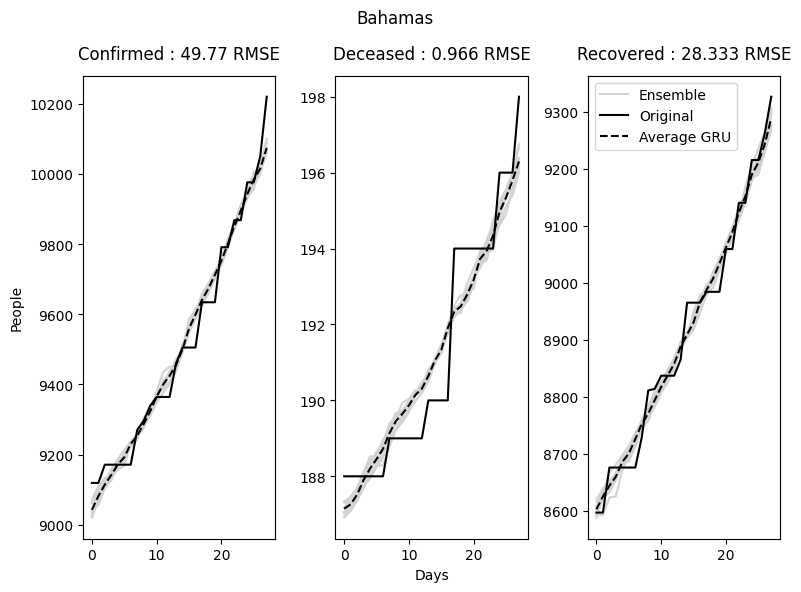}
        \caption{Bahamas}
    \end{subfigure}
    }
    \resizebox{\textwidth}{!}{
    \begin{subfigure}{0.5\textwidth}
        \centering
        \includegraphics[scale=0.25]{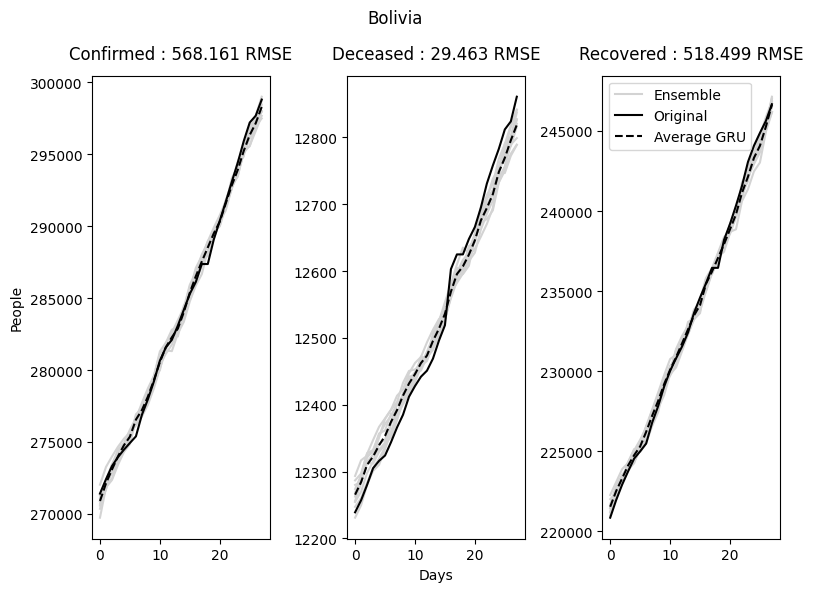}
        \caption{Bolivia}
    \end{subfigure}%
    \begin{subfigure}{0.5\textwidth}
        \centering
        \includegraphics[scale=0.25]{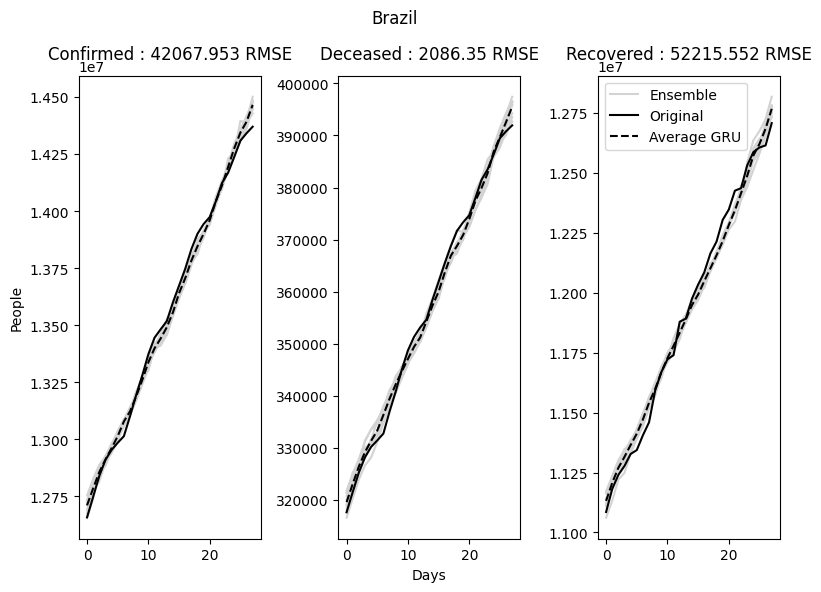}
        \caption{Brazil}
    \end{subfigure}
    }
    \resizebox{\textwidth}{!}{
    \begin{subfigure}{0.5\textwidth}
        \centering
        \includegraphics[scale=0.25]{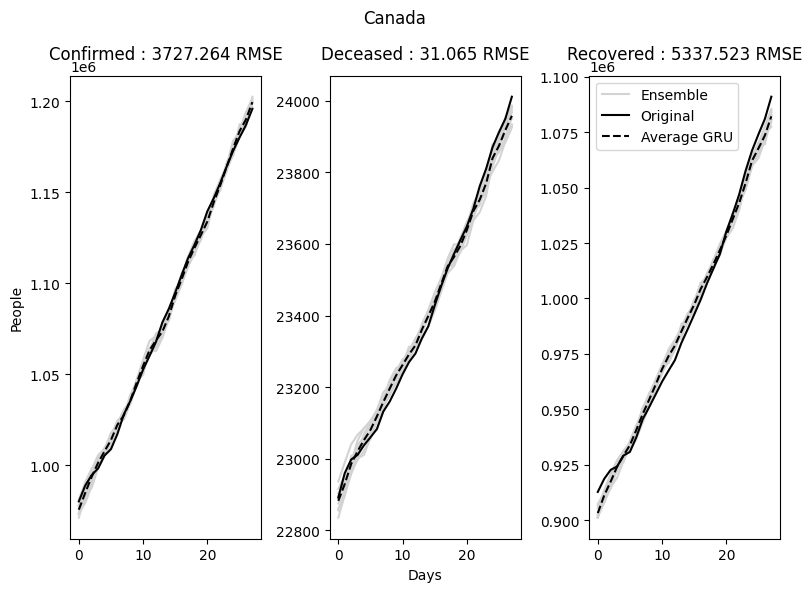}
        \caption{Canada}
    \end{subfigure}%
    \begin{subfigure}{0.5\textwidth}
        \centering
        \includegraphics[scale=0.25]{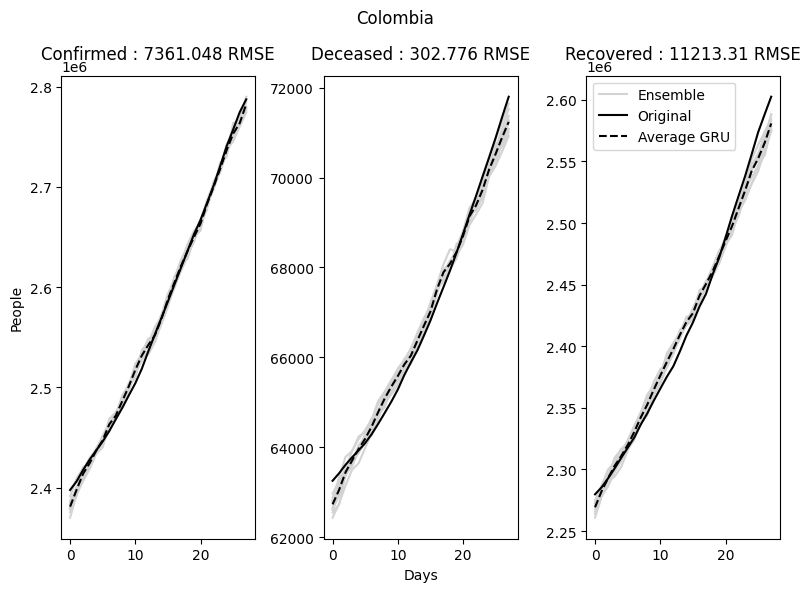}
        \caption{Colombia}
        \label{fig:Colombia}
    \end{subfigure}
    }
    \resizebox{\textwidth}{!}{
    \begin{subfigure}{0.5\textwidth}
        \centering
        \includegraphics[scale=0.25]{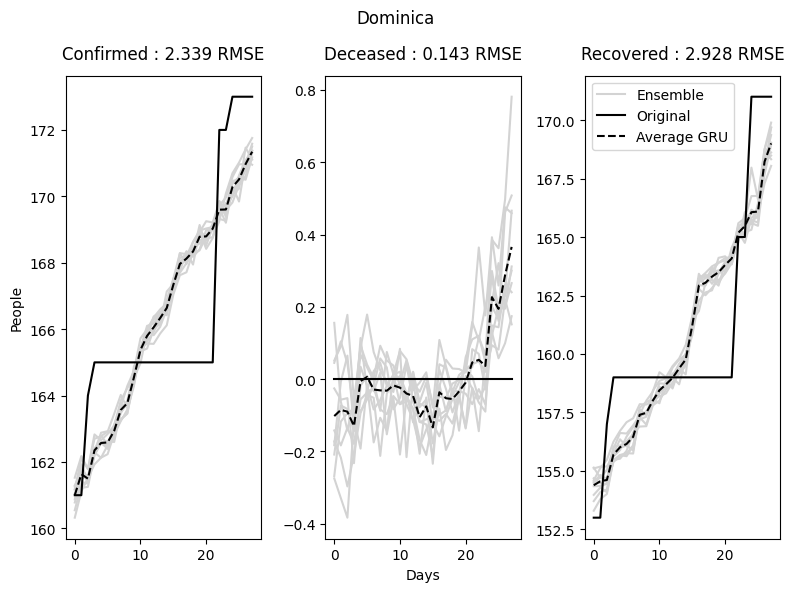}
        \caption{Dominica}
    \end{subfigure}%
    \begin{subfigure}{0.5\textwidth}
        \centering
        \includegraphics[scale=0.25]{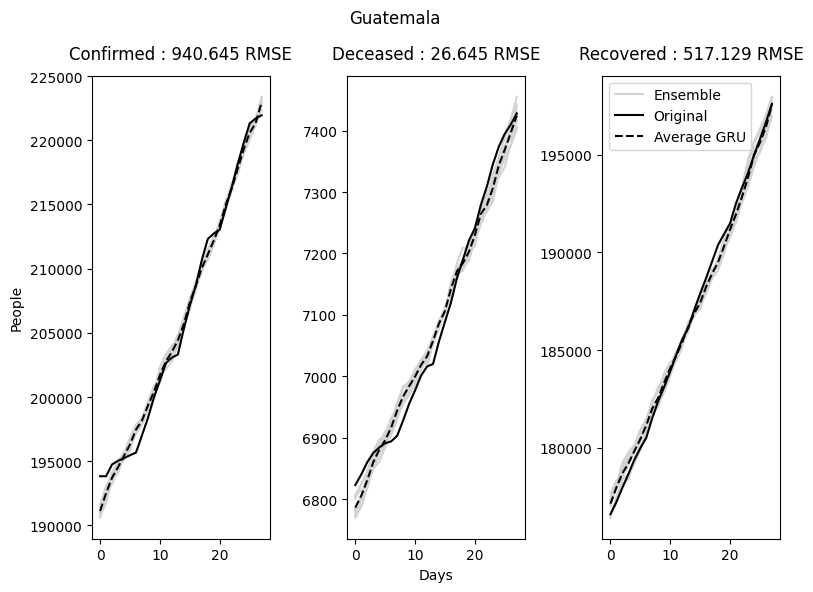}
        \caption{Guatemala}
    \end{subfigure}
    }
    \resizebox{\textwidth}{!}{
    \begin{subfigure}{0.5\textwidth}
        \centering
        \includegraphics[scale=0.25]{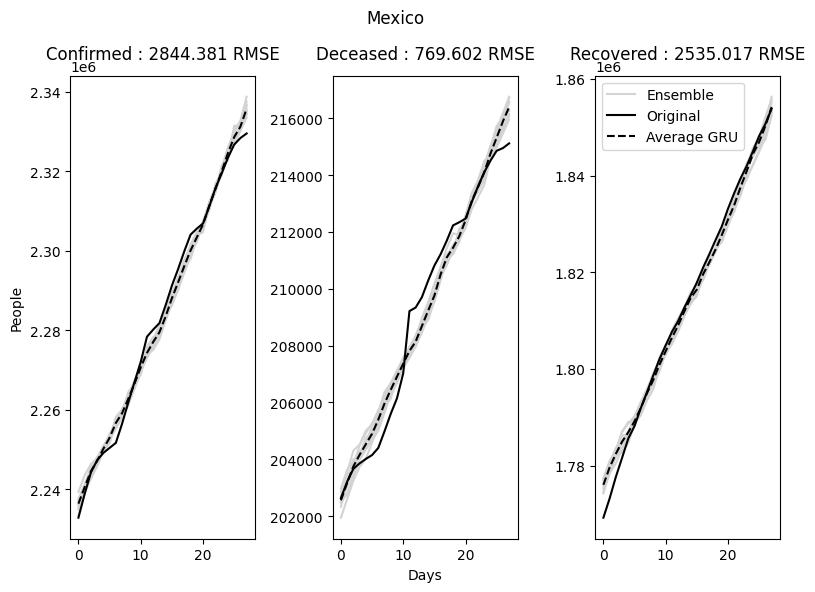}
        \caption{Mexico}
        \label{fig:Mexico}
    \end{subfigure}%
    \begin{subfigure}{0.5\textwidth}
        \centering
        \includegraphics[scale=0.25]{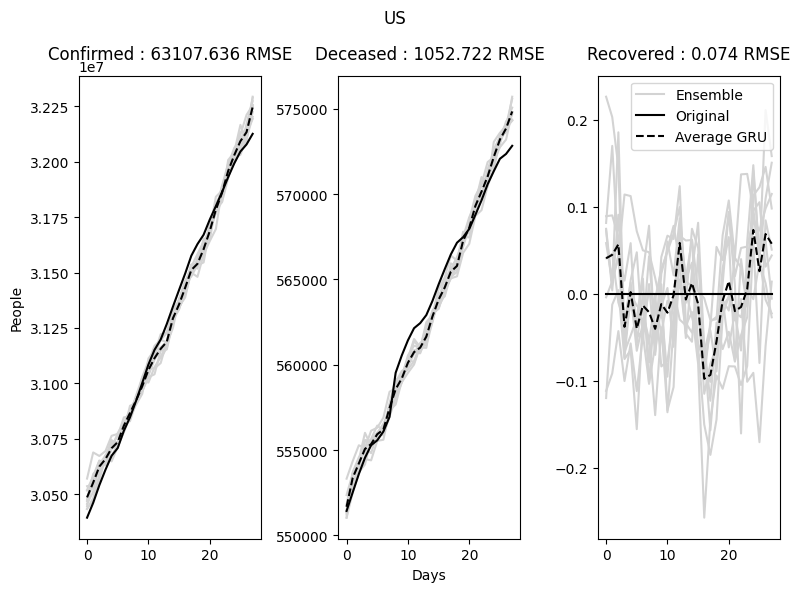}
        \caption{US}
        \label{fig:US}
    \end{subfigure}
    }
    \caption{America form 30/03/2021 to 27/04/2021}
    \label{fig:America}
\end{figure*}
\begin{figure*}
    \centering
    
    \resizebox{\textwidth}{!}{
    \begin{subfigure}{0.5\textwidth}
        \centering
        \includegraphics[scale=0.25]{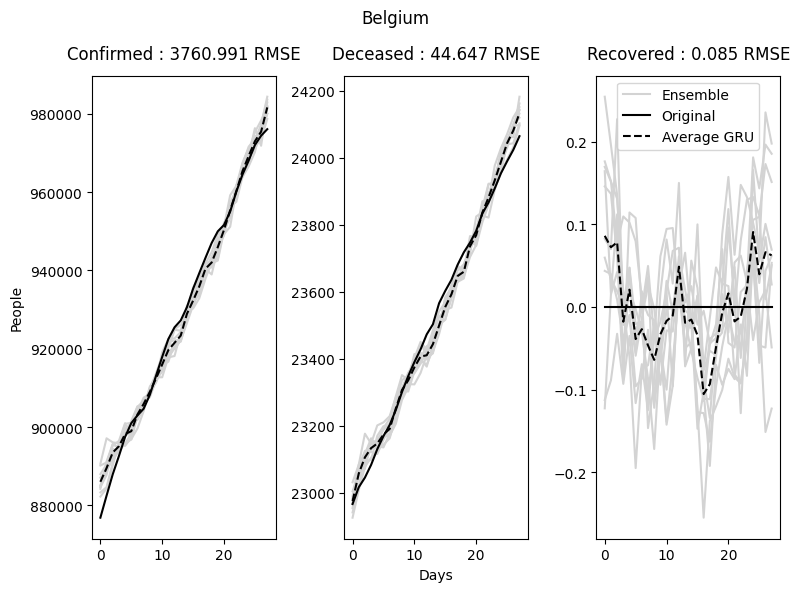}
        \caption{Belgium}
        \label{fig:Belgium}
    \end{subfigure}%
    \begin{subfigure}{0.5\textwidth}
        \centering
        \includegraphics[scale=0.25]{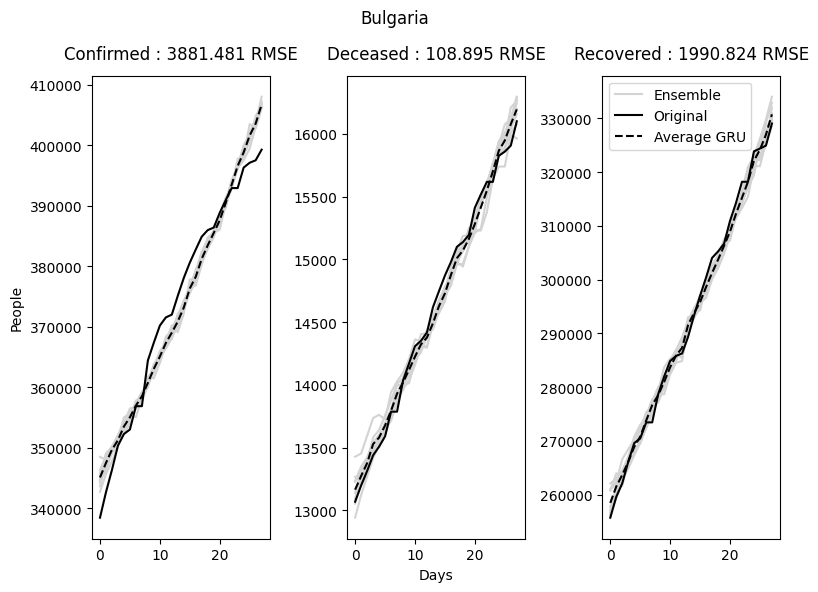}
        \caption{Bulgaria}
    \end{subfigure}
    }
    \resizebox{\textwidth}{!}{
    \begin{subfigure}{0.5\textwidth}
        \centering
        \includegraphics[scale=0.25]{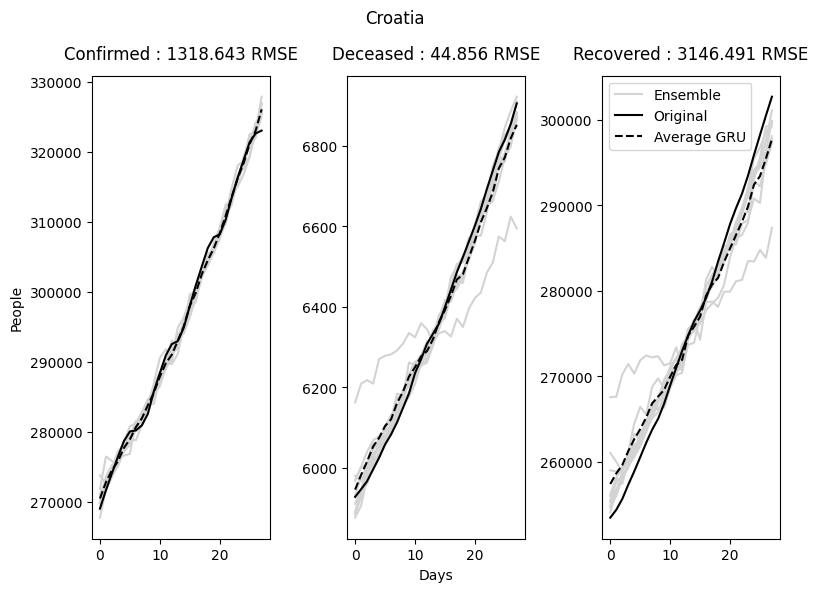}
        \caption{Croatia}
    \end{subfigure}%
    \begin{subfigure}{0.5\textwidth}
        \centering
        \includegraphics[scale=0.25]{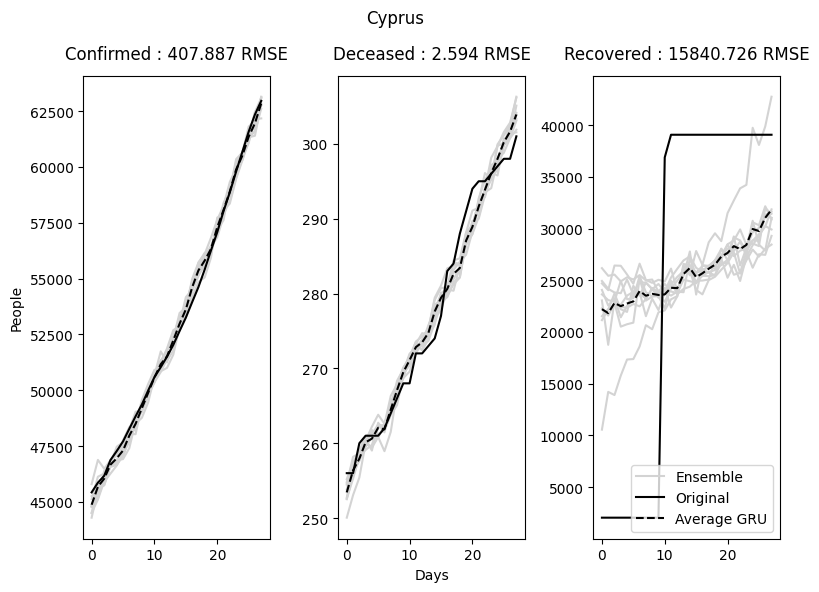}
        \caption{Cyprus}
        \label{fig:Cyprus}
    \end{subfigure}
    }
    \resizebox{\textwidth}{!}{
    \begin{subfigure}{0.5\textwidth}
        \centering
        \includegraphics[scale=0.25]{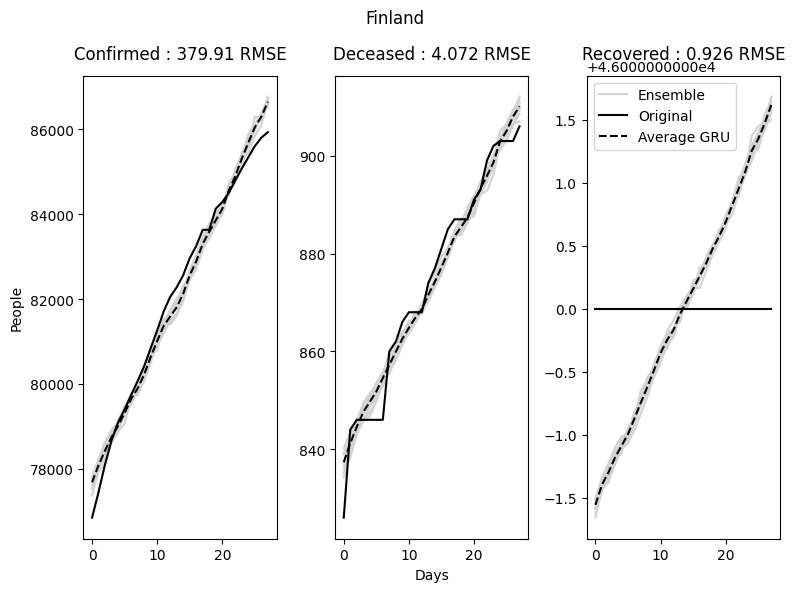}
        \caption{Finland}
    \end{subfigure}%
    \begin{subfigure}{0.5\textwidth}
        \centering
        \includegraphics[scale=0.25]{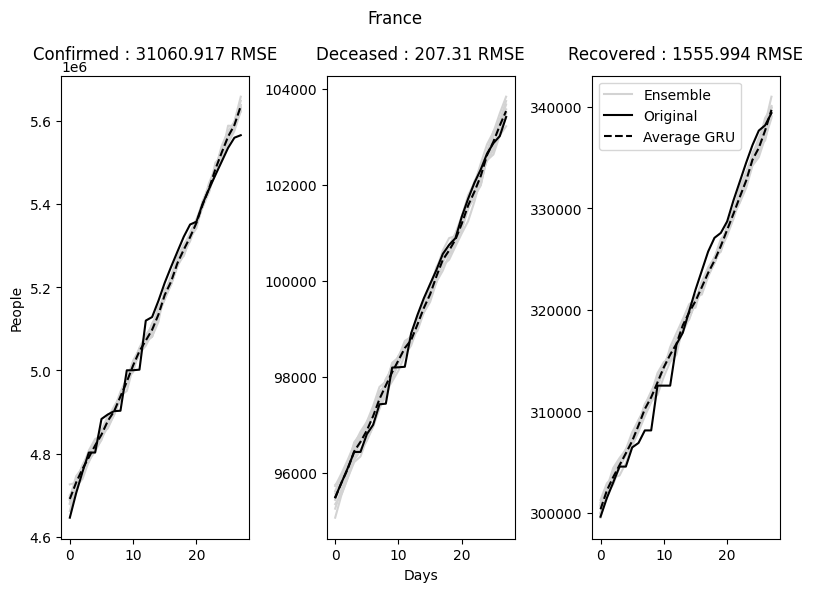}
        \caption{France}
    \end{subfigure}
    }
    \resizebox{\textwidth}{!}{
    \begin{subfigure}{0.5\textwidth}
        \centering
        \includegraphics[scale=0.25]{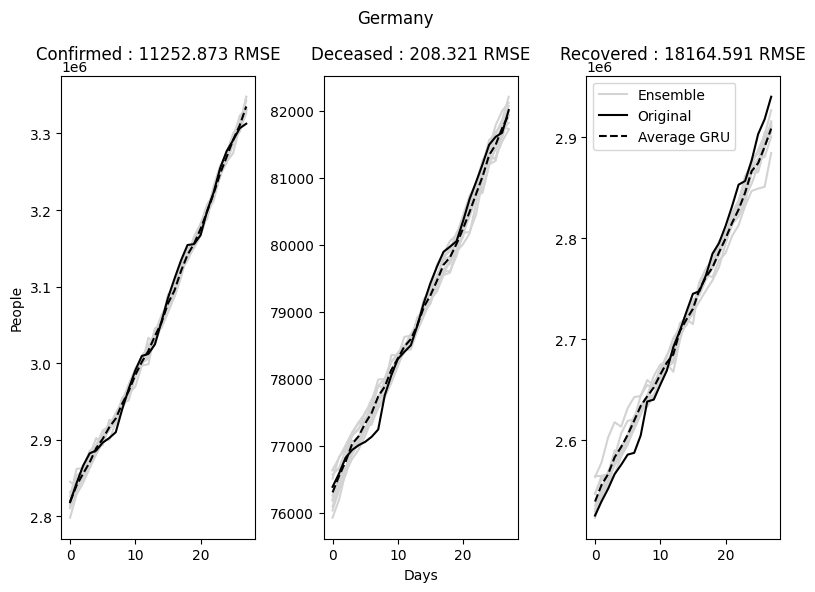}
        \caption{Germany}
    \end{subfigure}%
    \begin{subfigure}{0.5\textwidth}
        \centering
        \includegraphics[scale=0.25]{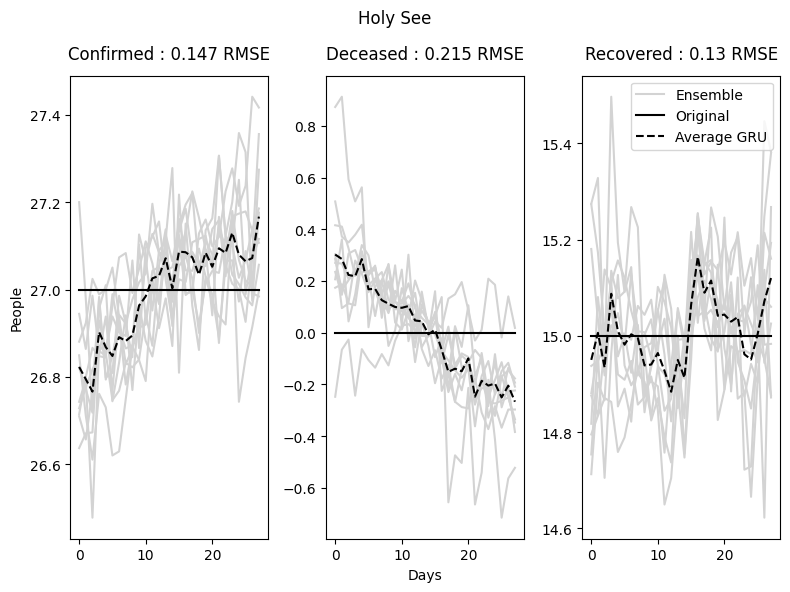}
        \caption{Holy See}
        \label{fig:Holy_See}
    \end{subfigure}
    }
    \resizebox{\textwidth}{!}{
    \begin{subfigure}{0.5\textwidth}
        \centering
        \includegraphics[scale=0.25]{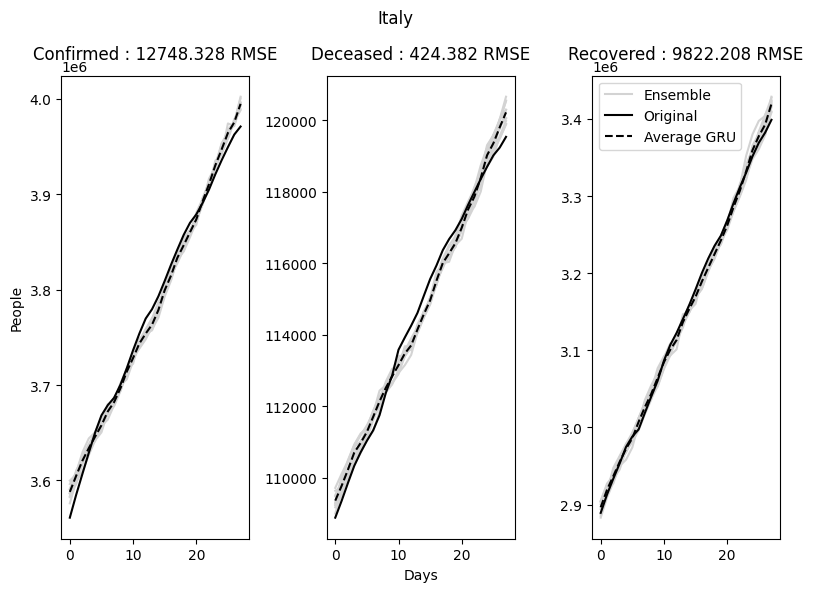}
        \caption{Italy}
    \end{subfigure}%
    \begin{subfigure}{0.5\textwidth}
        \centering
        \includegraphics[scale=0.25]{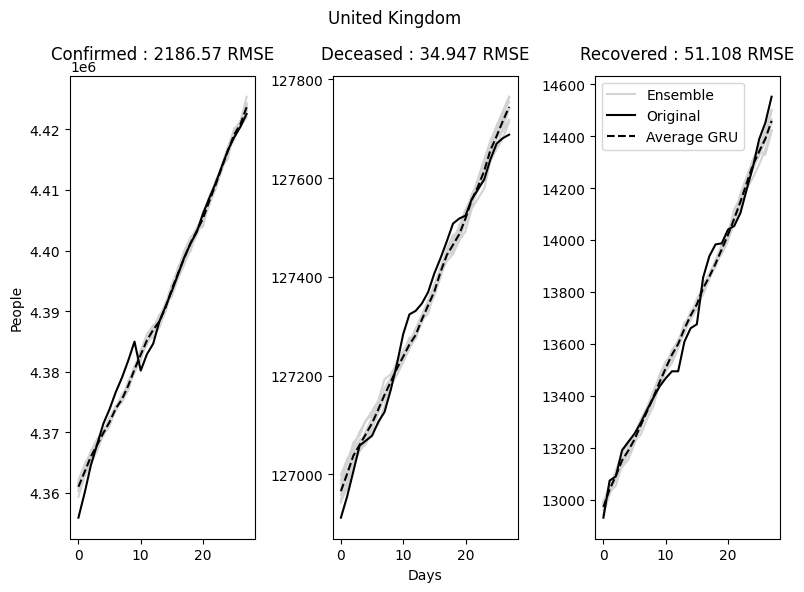}
        \caption{United Kingdom}
    \end{subfigure}
    }
    \caption{Europe form 30/03/2021 to 27/04/2021}
    \label{fig:Europe}
\end{figure*}
\begin{figure*}
    \centering
    
    \resizebox{\textwidth}{!}{
    \begin{subfigure}{0.5\textwidth}
        \centering
        \includegraphics[scale=0.25]{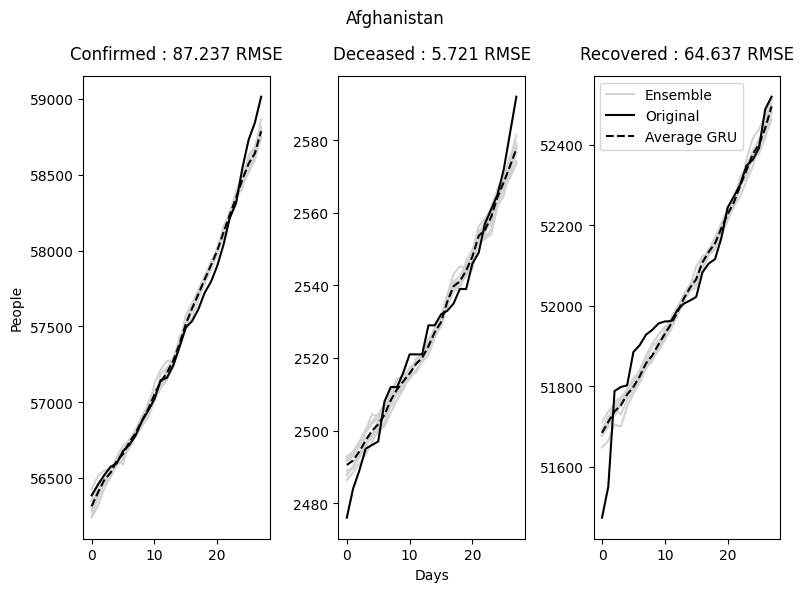}
        \caption{Afghanistan}
    \end{subfigure}%
    \begin{subfigure}{0.5\textwidth}
        \centering
        \includegraphics[scale=0.25]{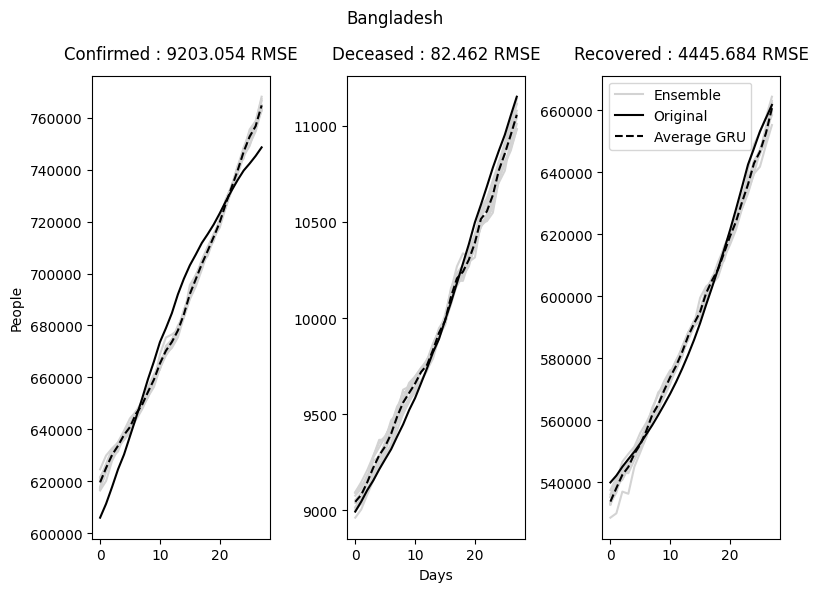}
        \caption{Bangladesh}
    \end{subfigure}
    }
    \resizebox{\textwidth}{!}{
    \begin{subfigure}{0.5\textwidth}
        \centering
        \includegraphics[scale=0.25]{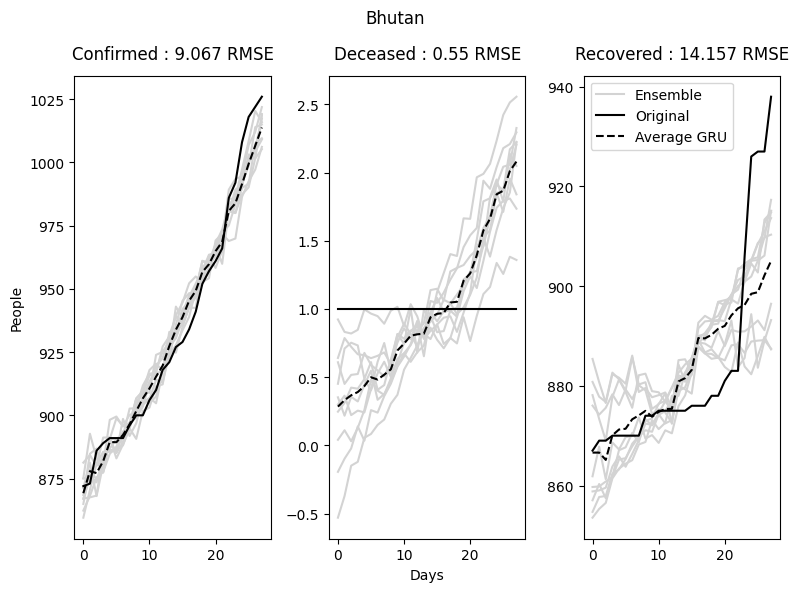}
        \caption{Bhutan}
        \label{fig:Bhutan}
    \end{subfigure}%
    \begin{subfigure}{0.5\textwidth}
        \centering
        \includegraphics[scale=0.25]{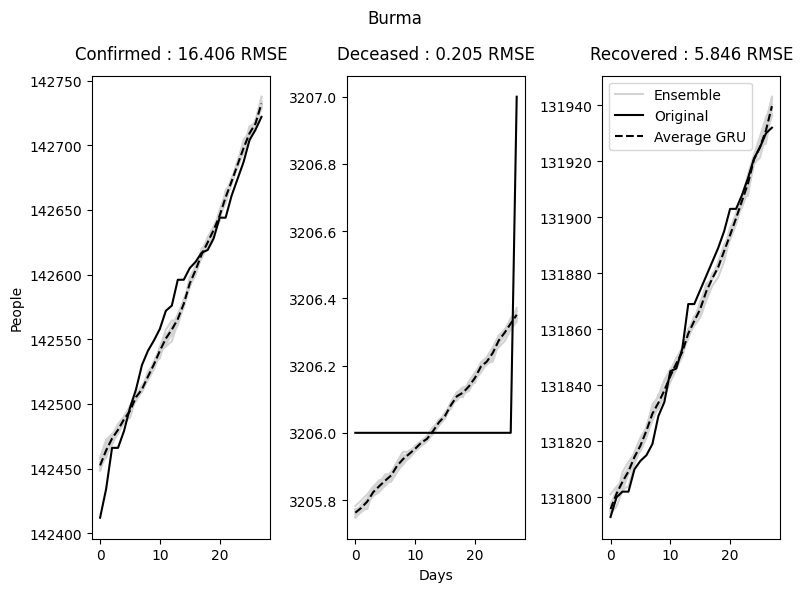}
        \caption{Burma}
    \end{subfigure}
    }
    \resizebox{\textwidth}{!}{
    \begin{subfigure}{0.5\textwidth}
        \centering
        \includegraphics[scale=0.25]{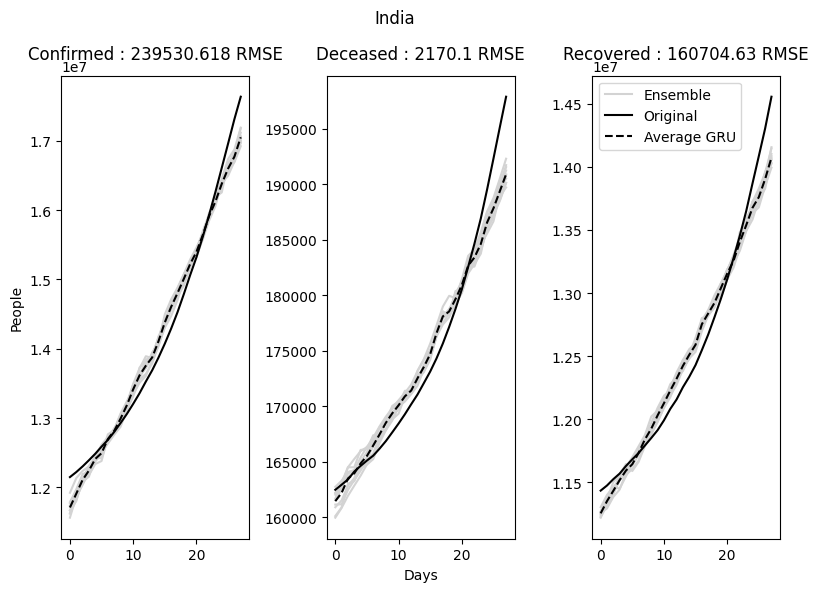}
        \caption{India}
    \end{subfigure}%
    \begin{subfigure}{0.5\textwidth}
        \centering
        \includegraphics[scale=0.25]{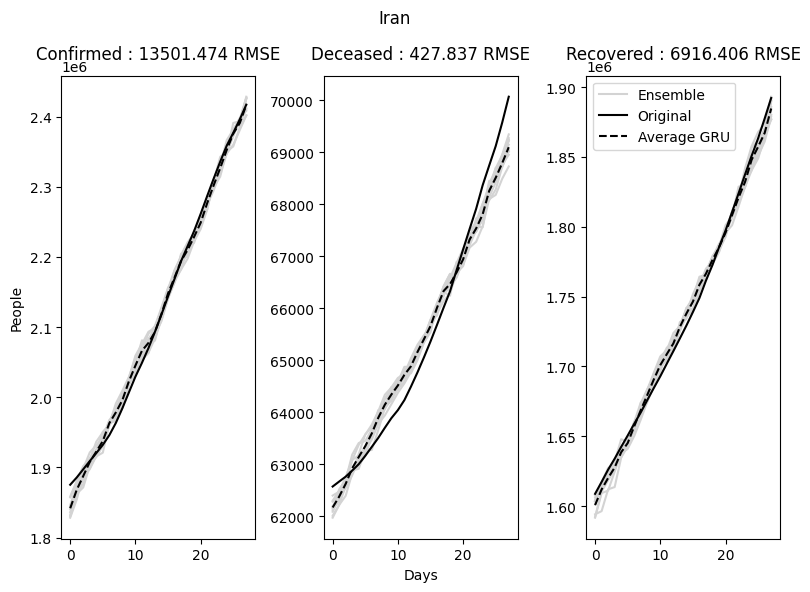}
        \caption{Iran}
    \end{subfigure}
    }
    \resizebox{\textwidth}{!}{
    \begin{subfigure}{0.5\textwidth}
        \centering
        \includegraphics[scale=0.25]{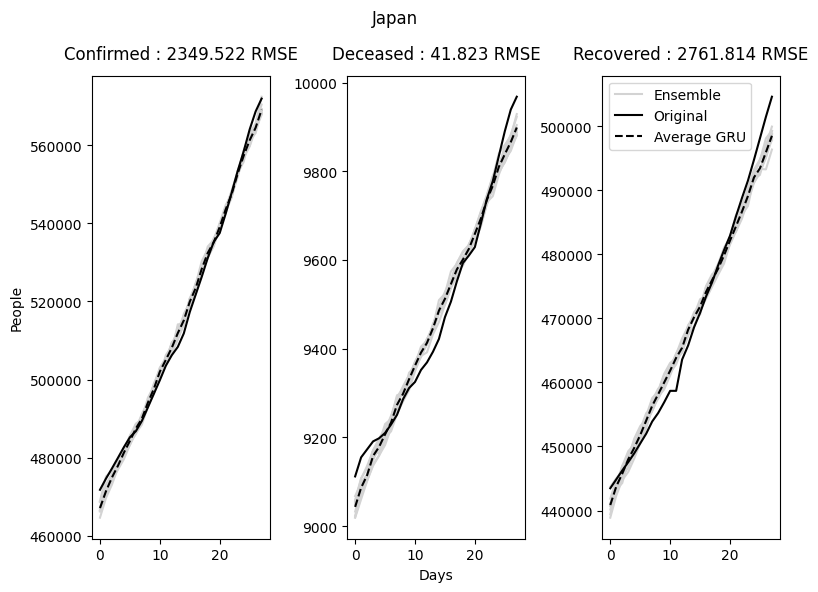}
        \caption{Japan}
        \label{fig:Japan}
    \end{subfigure}%
    \begin{subfigure}{0.5\textwidth}
        \centering
        \includegraphics[scale=0.25]{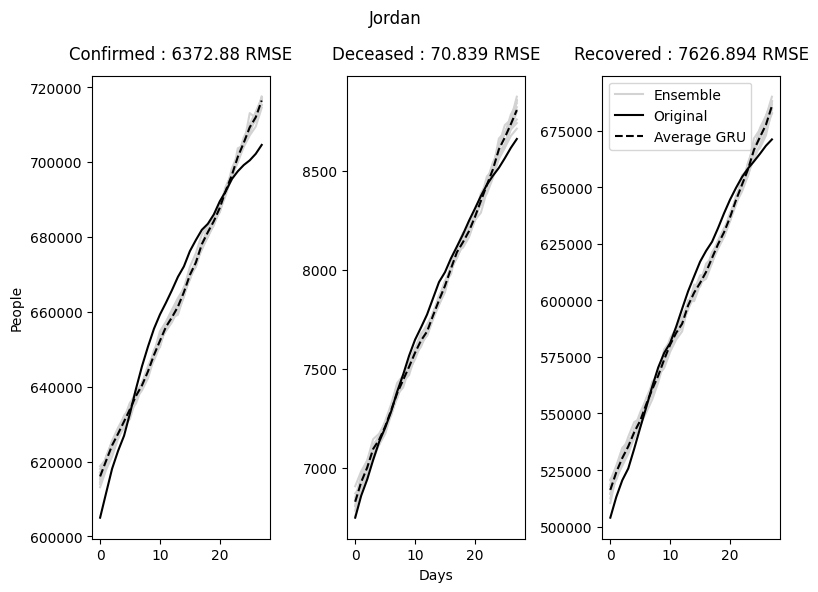}
        \caption{Jordan}
        \label{fig:Jordan}
    \end{subfigure}
    }
    \resizebox{\textwidth}{!}{
    \begin{subfigure}{0.5\textwidth}
        \centering
        \includegraphics[scale=0.25]{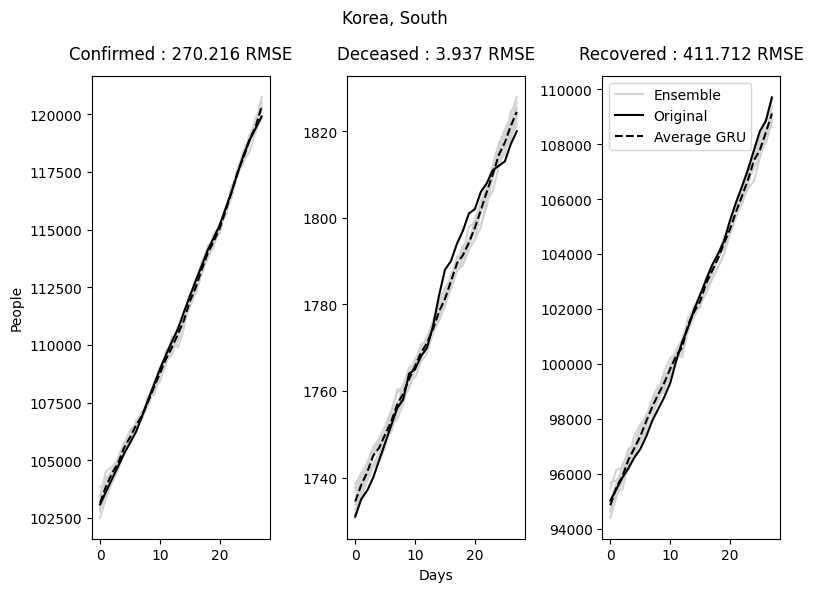}
        \caption{South Korea}
    \end{subfigure}%
    \begin{subfigure}{0.5\textwidth}
        \centering
        \includegraphics[scale=0.25]{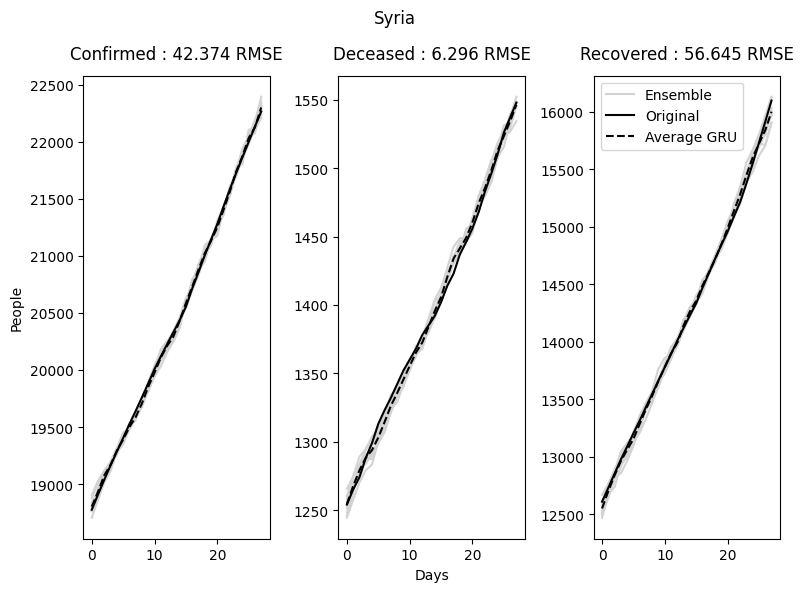}
        \caption{Syria}
    \end{subfigure}
    }
    \caption{Asia form 30/03/2021 to 27/04/2021}
    \label{fig:Asia}
\end{figure*}

\begin{figure*}
    \centering
    
    \resizebox{\textwidth}{!}{
    \begin{subfigure}{0.5\textwidth}
        \centering
        \includegraphics[scale=0.25]{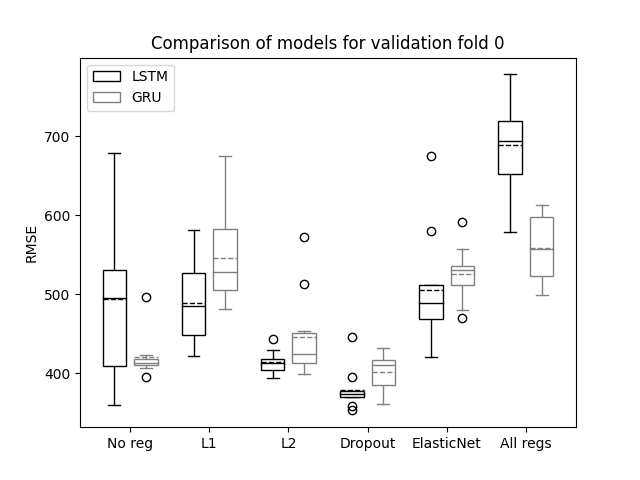}
    \end{subfigure}%
    \begin{subfigure}{0.5\textwidth}
        \centering
        \includegraphics[scale=0.25]{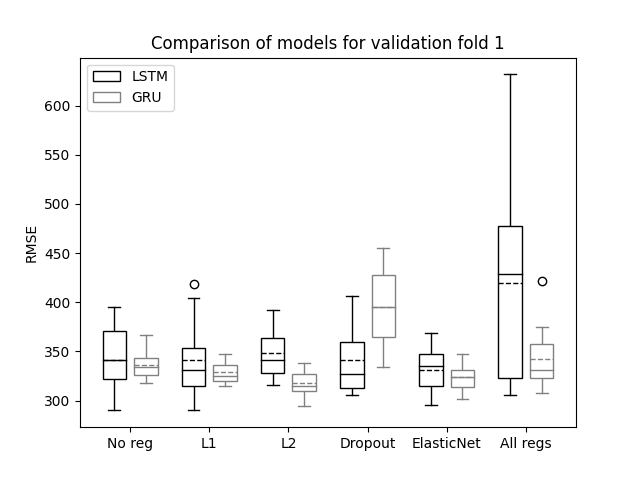}
    \end{subfigure}
    }
    \resizebox{\textwidth}{!}{
    \begin{subfigure}{0.5\textwidth}
        \centering
        \includegraphics[scale=0.25]{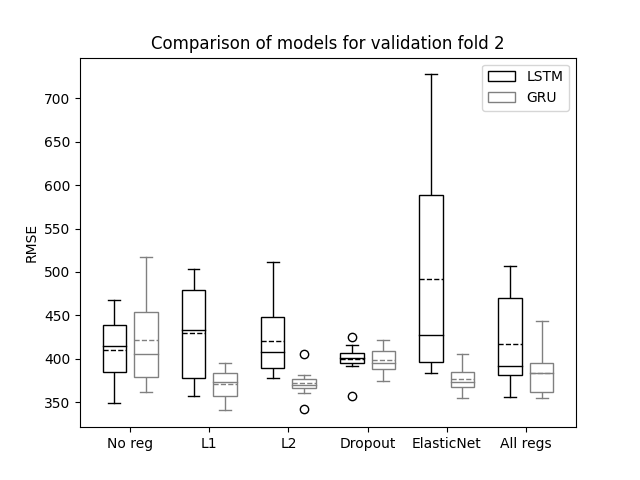}
    \end{subfigure}%
    \begin{subfigure}{0.5\textwidth}
        \centering
        \includegraphics[scale=0.25]{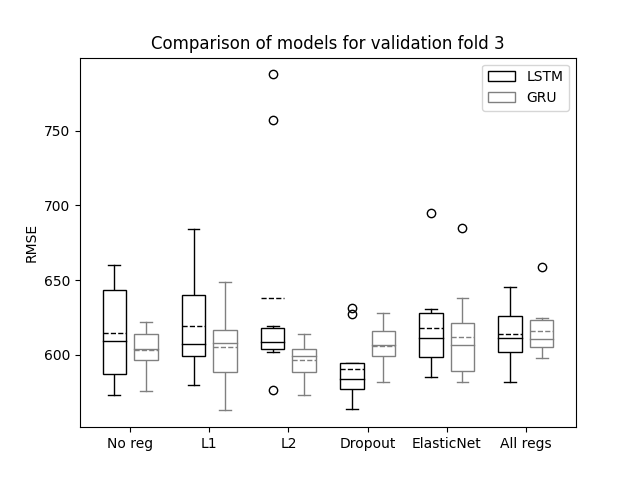}
    \end{subfigure}
    }
    \resizebox{\textwidth}{!}{
    \begin{subfigure}{0.5\textwidth}
        \centering
        \includegraphics[scale=0.25]{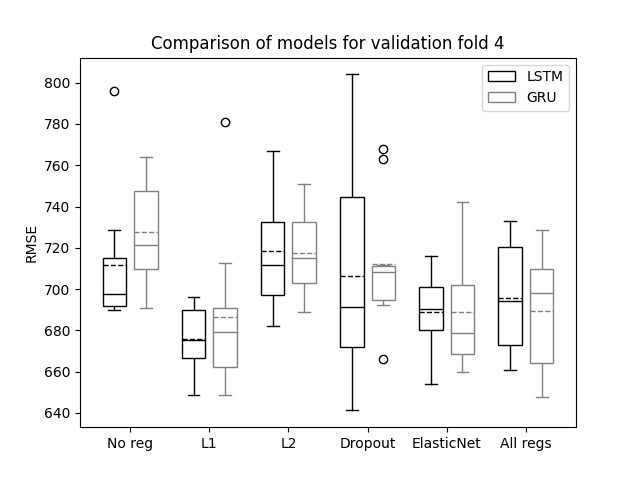}
    \end{subfigure}%
    \begin{subfigure}{0.5\textwidth}
        \centering
        \includegraphics[scale=0.25]{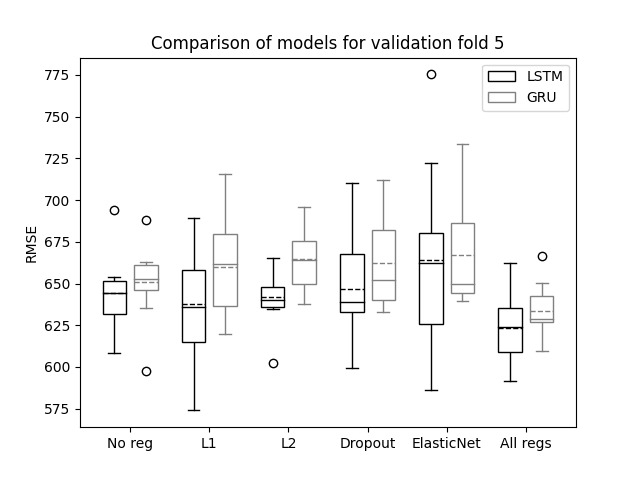}
    \end{subfigure}
    }
    \resizebox{\textwidth}{!}{
    \begin{subfigure}{0.5\textwidth}
        \centering
        \includegraphics[scale=0.25]{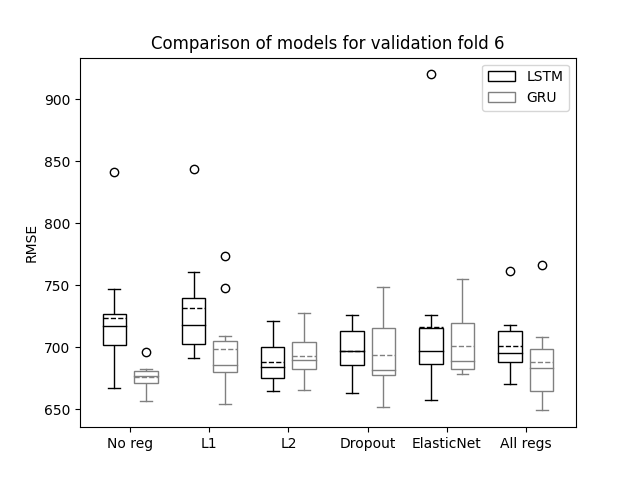}
    \end{subfigure}%
    \begin{subfigure}{0.5\textwidth}
        \centering
        \includegraphics[scale=0.25]{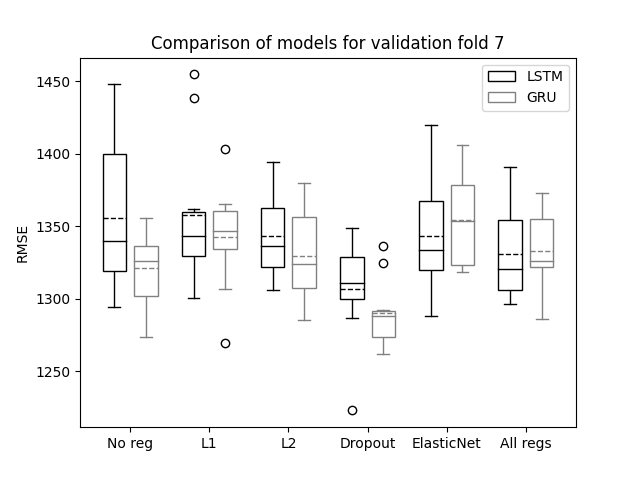}
    \end{subfigure}
    }
    \resizebox{\textwidth}{!}{
    \begin{subfigure}{0.5\textwidth}
        \centering
        \includegraphics[scale=0.25]{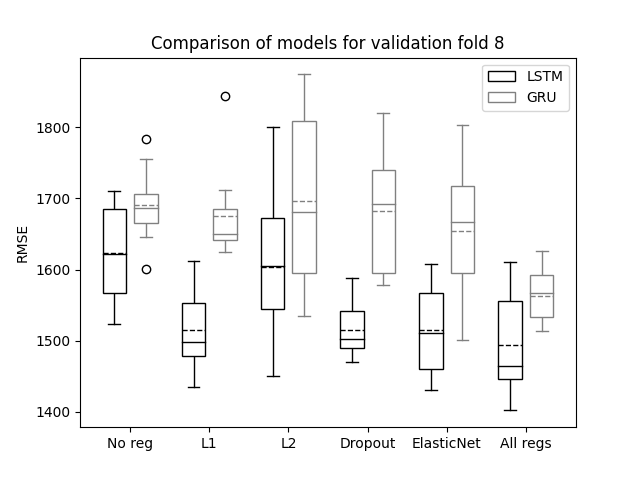}
    \end{subfigure}%
    \begin{subfigure}{0.5\textwidth}
        \centering
        \includegraphics[scale=0.25]{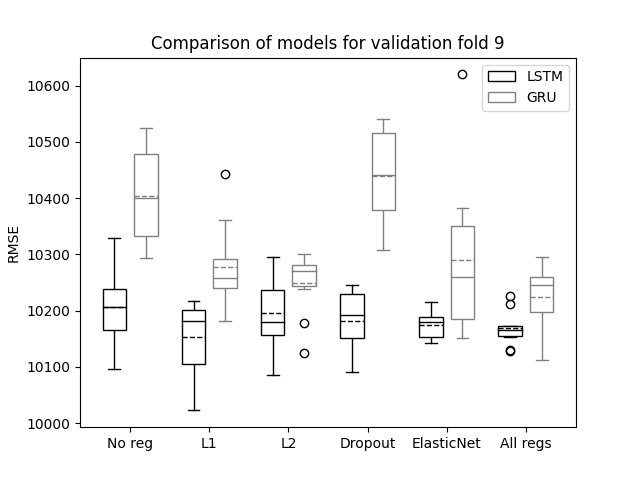}
    \end{subfigure}
    }
    \resizebox{\textwidth}{!}{
    \begin{subfigure}{0.5\textwidth}
        \centering
        \includegraphics[scale=0.25]{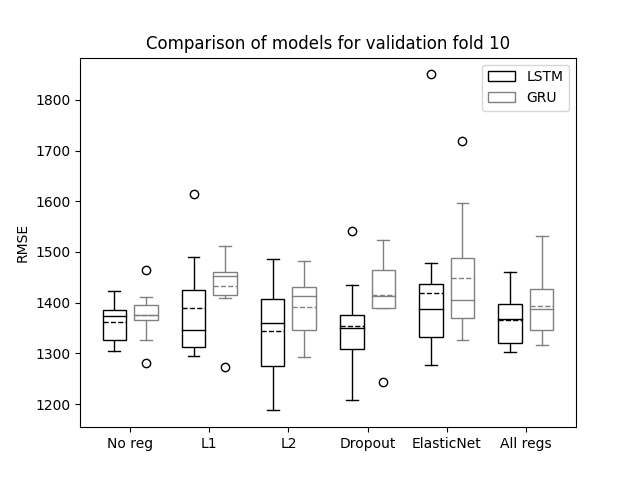}
    \end{subfigure}%
    \begin{subfigure}{0.5\textwidth}
        \centering
        \includegraphics[scale=0.25]{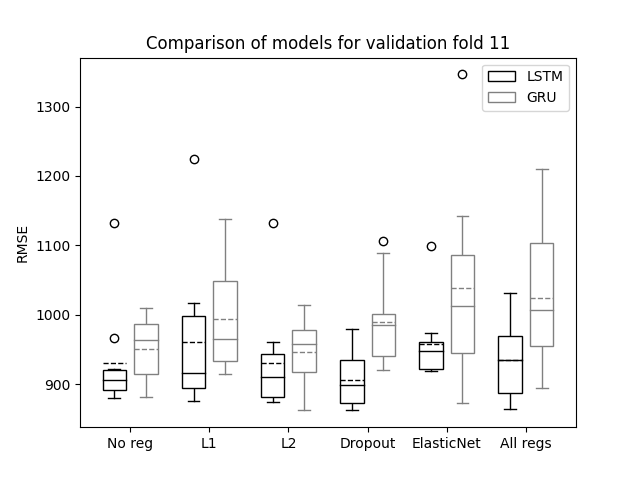}
    \end{subfigure}
    }
    \resizebox{\textwidth}{!}{
    \begin{subfigure}{0.5\textwidth}
        \centering
        \includegraphics[scale=0.25]{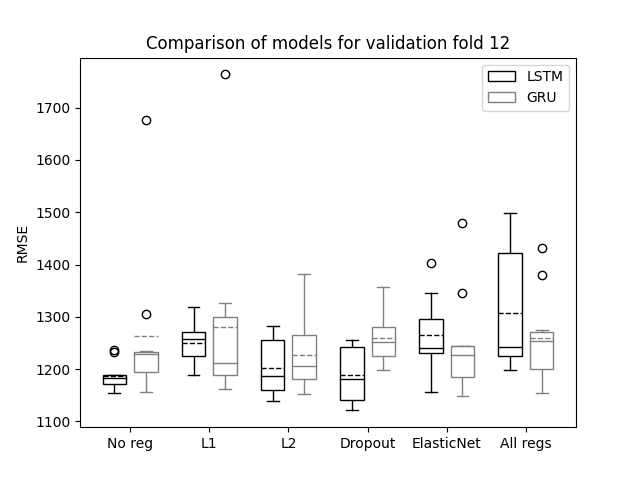}
    \end{subfigure}%
    \begin{subfigure}{0.5\textwidth}
        \centering
        \includegraphics[scale=0.25]{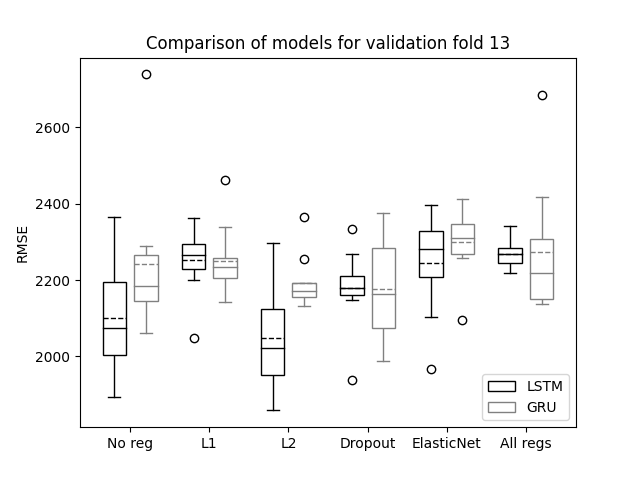}
    \end{subfigure}
    }
    \caption{Box plots for all validation folds.}
\end{figure*}

\bibliographystyle{plainnat}
\bibliography{COVID-19_research-paper}

\end{document}